**Title:** What Don't You Understand? Using Large Language Models to Identify and Characterize Student Misconceptions About Challenging Topics

**Author information:**
Michael J. Parker, MD was, at the time this work was conducted, the associate dean for online learning research, faculty director of HMX, Office of External Education, and assistant professor of medicine, Harvard Medical School and Division of Pulmonary, Critical Care and Sleep Medicine, Beth Israel Deaconess Medical Center, Boston, Massachusetts; ORCID: https://orcid.org/0000-0003-4739-5217, email: mparker@alum.mit.edu.

María G. Zavala-Cerna
maria.cerna@edu.uag.mx
Decanato Medicina
Universidad Autónoma de Guadalajara
ORCID 0000-0001-6508-9657

**Correspondence should be addressed to** Michael J. Parker; email: mparker@alum.mit.edu.

**Acknowledgements:** The authors would like to thank the entire HMX team for their contributions to creating and administering the courses and the UAG faculty for their feedback on uses for the challenging topics information.

**Funding statement:**
This research received no external funding.

**Competing interests:**
The authors have no relevant financial or non-financial interests to disclose.

**Ethics approval and consent to participate:**
The study was conducted using anonymized data after the completion of the courses and was determined not to be human subjects research by the HMS Office of Human Research Administration and the UAG Institutional Review Board.

**Availability of data and materials:**
The prompts are shared in the appendix. To help preserve the integrity of the courses, the quiz questions are not included in an open-access repository. Other data may be provided upon request to the authors and approval of the university research ethics board.

**Title:** What Don’t You Understand? Using Large Language Models to Identify and Characterize Student Misconceptions About Challenging Topics

# Abstract

This study presents a systematic approach to identifying and characterizing student misconceptions in online learning environments through a novel combination of quantitative performance analysis and large language model (LLM) assessment. We analyzed data from 9 course periods across 5 online biomedical science courses, encompassing 3,802 medical student enrollments. Using data from 40-50 topic-focused quizzes per course, we developed a two-stage methodology. First, we identified challenging central topics using quiz-level performance metrics. Second, we employed LLMs to characterize the underlying misconceptions in these high-priority areas. By examining student performance on first attempts across primarily multiple-choice questions (MCQs), we identified consistently challenging topics that were also central to course objectives. We then leveraged recent advances in generative AI to analyze three distinct data sources in combination: quiz question content, student response patterns, and lecture transcripts. This approach revealed actionable insights about student misconceptions that were not apparent from performance data alone. The quality of the LLM-identified misconceptions was rated as excellent by subject matter experts. We also conducted teacher interviews to assess the perceived utility of our topic identification method. Faculty found that data-driven identification of challenging topics was valuable and corroborated their own classroom observations. This methodology provides a scalable approach to characterizing student difficulties in learning environments where quizzes are used. Our findings demonstrate the potential for targeted and potentially personalized interventions in future course iterations, with clear pathways for measuring intervention effectiveness through follow-up quiz performance.

# 1 Introduction

Identifying and characterizing challenging topics is a fundamental aspect of effective teaching and a key focus of educational research (Kennedy & Kraemer, 2018; Shulman, 1986; Slotta & Chi, 2006; Wiegand et al., 2021). Understanding *where* students struggle allows educators to prioritize content, while understanding *why* they struggle is critical for developing effective teaching approaches (Kaczmarczyk et al., 2010; Larkin, 2012). Insight into both the location and nature of these difficulties enables educators to strategically allocate resources and provide targeted support to enhance learning outcomes (Modell & Wenderoth, 2005).

The rapid growth of online and hybrid learning environments presents both challenges and opportunities for understanding student difficulties. Traditional methods for gauging student understanding, such as direct observation and immediate feedback (e.g., noticing confused expressions, fielding clarifying questions), are often limited in digital spaces. In contrast, these online environments generate rich data that can be systematically analyzed to identify patterns of student struggle and success (Almeida & Monteiro, 2021; Kearns, 2012; Quadri & Shukor, 2021; Ramos & Castillo, 2024).

The field of learning analytics offers promising avenues for identifying student challenges (Ferguson, 2012; Nunn et al., 2016; Wise, 2019). Integrating learning analytics with formative assessment data, particularly topic-focused quizzes, provides a unique opportunity to develop more nuanced and responsive educational strategies. Unlike summative assessments that evaluate learning at the end of a unit, formative assessments are embedded throughout the learning process, providing continuous insights into student understanding (Black & Wiliam, 1998; Mustamin, 2024). In this context, we operationally define a 'struggle' not as a momentary lapse, but as a persistent inability to demonstrate mastery of a specific topic on a first attempt. For example, if a formative quiz on immune signaling consistently yields low scores across cohorts, it reveals a 'struggle' localized to that concept. Unlike a final exam which identifies failure after the fact, this real-time data highlights where students face cognitive hurdles while learning is still occurring. Analyzing data from these assessments across multiple cohorts allows educators to systematically identify consistently challenging topics. This allows for iterative curriculum improvements and targeted interventions.

In contrast to individual question analysis, which may be confounded by question-specific issues such as ambiguous wording or inadequate coverage in course materials, quiz-level analysis can reveal broader conceptual challenges that persist across multiple questions within a topic area (Cook & Beckman, 2006; Glaser et al., 2001; Haladyna & Downing, 2011). This approach allows for a more robust identification of fundamental learning difficulties while minimizing the impact of question-specific confounds.

While identifying difficult topics is a crucial step, understanding the underlying misconceptions that lead to these difficulties is equally important for developing effective interventions (Posner et al., 1982; Verkade et al., 2018; Wandersee et al., 1994). Misconceptions are incorrect understandings that actively interfere with new learning. These generally fall into two categories: factual misconceptions, which involve incorrect recall of specific details (e.g., memorizing the wrong name of an enzyme), and conceptual misconceptions, which involve flawed mental models of how a system functions (e.g., misunderstanding the cause-and-effect relationship in a pathway) (Suprapto, 2020). This study focuses primarily on conceptual misconceptions. Rather than viewing any incorrect answer as a misconception, we identify them by analyzing the most frequently selected distractors within the consistently challenging topics identified by our performance analysis. This ensures we are isolating shared, underlying flaws in reasoning rather than identifying artifacts of isolated, poorly worded questions. It is also important to distinguish these misconceptions from 'bottlenecks' or difficulties with 'threshold concepts,' where students may struggle due to the inherent complexity or abstract nature of new material without necessarily holding a false belief (Meyer & Land, 2003). While our performance analysis identifies difficult topics that may be bottlenecks, our LLM analysis specifically seeks to isolate the misconceptions – the active misunderstandings – that often contribute to that difficulty. Gaining insight into students' underlying reasons for difficulty allows teachers to develop instructional approaches that directly address these fundamental areas where students may have misconceptions or gaps in their understanding, rather than simply re-teaching the material in the same way.

Educators have predominantly used qualitative methods to derive misconceptions based on student responses or wrong answers (Coll & Treagust, 2003; Gurel et al., 2015; Halim et al., 2018; Hestenes et al., 1992; Treagust, 1988). These methods, such as think-aloud protocols, interviews, and concept mapping, provide rich insights but are time-consuming and difficult to scale to large classes, particularly in online environments (Olde Bekkink et al., 2016). Prior researchers have utilized natural language processing to identify misconceptions in open-ended text and programming problems. However, these techniques, predating generative AI, often relied on embeddings and clustering, still necessitating significant expert interpretation to identify common misconceptions within the data (Gorgun & Botelho, 2023; Hoq et al., 2024; Kökver et al., 2025; Lo et al., 2021; Michalenko et al., 2017; Shi et al., 2021). Prior studies involve having experts view and interpret the entries to identify a common misconception among the submissions in each cluster. Therefore, despite automation of certain portions, these methods still require significant human oversight and interpretation.

The emergence of generative AI offers a potential path forward. Early work has begun to explore the use of large language models (LLMs) for understanding student thinking. As part of their research questions, Smart et al. (2024) investigated whether language models could interpret students' wrong answers on National Assessment of Educational Progress (grades 4, 8, and 12 math and science) items. They found that GPT-4's explanations of frequently chosen wrong answers corresponded well with those of experienced science teachers, highlighting the potential of LLMs for this task.

Building upon such initial explorations, recent advances in LLMs more broadly offer powerful new capabilities for analyzing and characterizing student misconceptions comprehensively and at scale. Moving beyond interpreting individual wrong answers, these models can potentially analyze multiple, diverse sources of educational data simultaneously – including quiz questions, student response patterns, and lecture transcripts – to identify underlying patterns of conceptual misunderstandings. This technological capability enables a systematic approach to misconception identification that was previously impractical, particularly in large-scale learning environments.

Several key advancements in LLMs make them particularly well-suited for this task:

- **Extended Context Windows:** The emergence of LLMs with significantly longer context windows (maximum input size) allows for the analysis of extensive textual data, such as entire lecture transcripts or sets of related quiz questions, in a single processing instance (Levy et al., 2024; Zhang et al., 2024). This is crucial for understanding the relationships between concepts presented in different parts of a course and how students might be (mis)connecting them. Prior limitations in context length meant that subtle connections, crucial for understanding misconceptions, were often missed.

- **Enhanced Subject Matter Expertise:** LLMs have demonstrated increasingly sophisticated understanding of complex subjects, including graduate-level biomedical science (M. Liu et al., 2024; Telenti et al. 2024; M. Liu et al., 2025). This improved domain knowledge is essential for accurately interpreting student responses and identifying deviations from accepted scientific understanding.
- **Reduced Computational Cost:** The decreasing cost of accessing and utilizing powerful LLMs via APIs has made it feasible to process large datasets of educational materials and student responses (Cottier et al., 2025). This allows for the analysis of entire courses and multiple cohorts, providing a more comprehensive view of student learning.
- **Multimodal Capabilities:** Emerging multimodal LLMs can analyze not only text but also images and other media like video and audio (Carolan et al., 2024). This capability is particularly relevant for courses that incorporate visual representations, such as diagrams, charts, or microscopic images, which are common in biomedical science. Misconceptions can often arise from misinterpreting visual information. Multimodal LLMs have traditionally struggled with charts and scientific diagrams but are improving in these capabilities (Wang et al., 2024; Xiao & Zhang, 2025).
- **Structured Data Output:** Modern LLMs can be prompted to produce output in structured formats (e.g., JavaScript object notation, called JSON), making it easier to integrate their analyses into learning analytics dashboards and other educational tools (Y. Liu et al., 2024). This also facilitates the automation of misconception identification and reporting.

- **Ease of Use via API:** The availability of LLMs through APIs allows for batch processing of tasks, enabling efficient analysis of large datasets. This is essential for scaling misconception analysis to entire courses and multiple cohorts.
- **In-Context Learning and Grounding:** Techniques like in-context learning (including few-shot prompting or formatting examples in the prompt) or grounding (adding reliable reference content like transcripts to the prompt) can significantly mitigate the risk of LLM "hallucinations" (generating factually incorrect or nonsensical information) by providing the model with specific reference material, examples of the desired task, and output format (Huang et al., 2025). This is crucial for ensuring the reliability of misconception analysis.

The ability to identify and understand student misconceptions is a critical step towards "closing the learning analytics loop" (Ferguson et al., 2023; Motz et al. 2023). This refers to the cyclical process of collecting data on student learning, analyzing that data to identify areas for improvement, implementing interventions based on the analysis, and then collecting new data to evaluate the effectiveness of those interventions. By providing a scalable method for identifying and characterizing misconceptions, LLMs can help educators more effectively close this loop and continuously improve their teaching. This overall cycle is shown in Fig. 1.

**Fig. 1** A framework for identifying and addressing student misconceptions. The process begins with data collection and analysis to pinpoint challenging topics. Next, underlying misconceptions are characterized, potentially using LLMs. This analysis informs two intervention pathways:

cohort-level interventions for broad improvements and individualized learning paths for tailored support. The effectiveness of these interventions is then assessed, providing new data to further refine the pedagogical process and reveal remaining or new misconceptions.

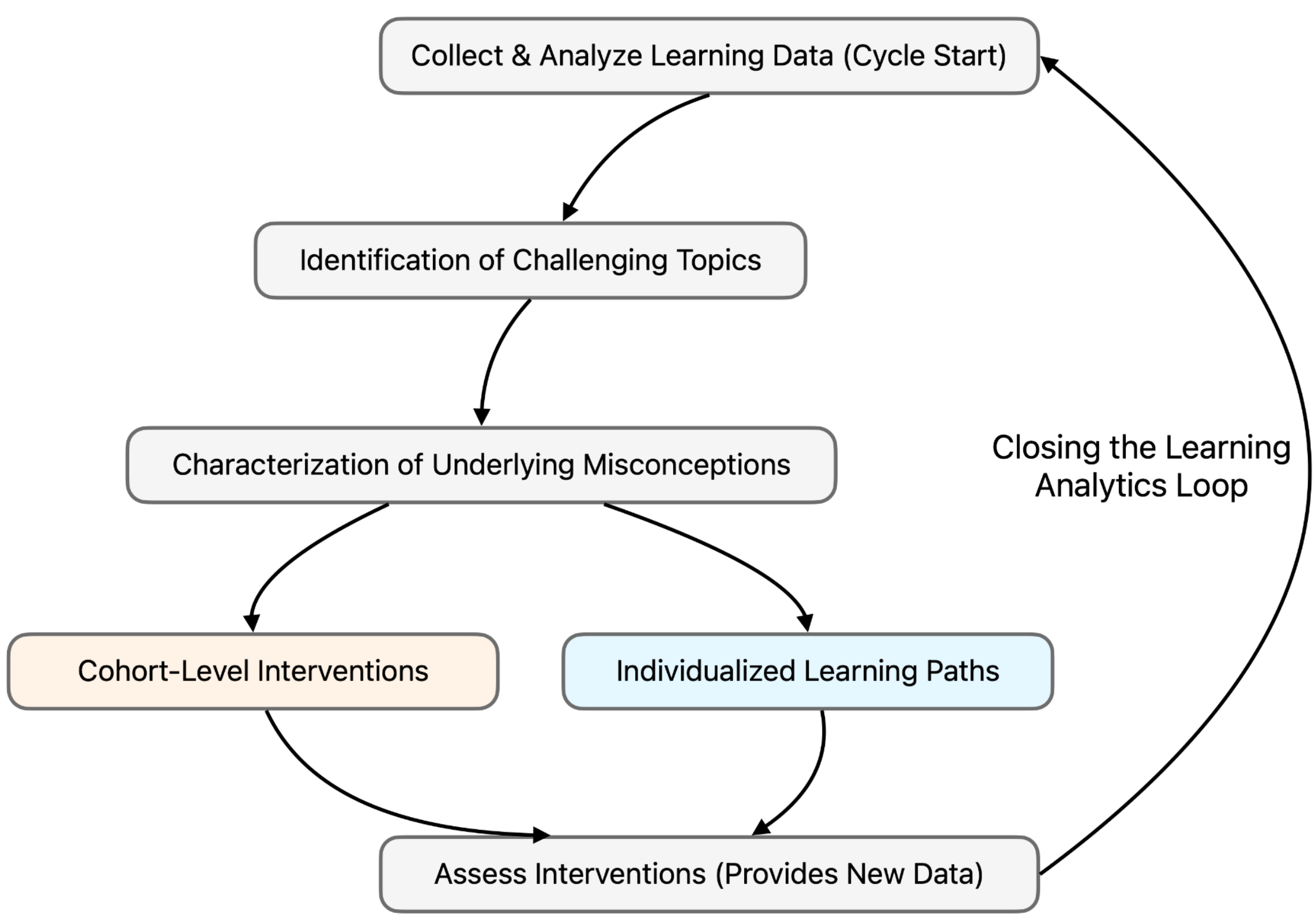

Furthermore, understanding misconceptions, whether at the cohort level or for individual students, can facilitate the design of more personalized interventions (Hattie & Timperley, 2007; Knight et al., 2014; Moskal et al., 2023). This aligns with the broader goals of adaptive learning, where instructional materials and activities are tailored to meet the specific needs and learning styles of each student (Chi et al., 2011). By identifying the specific misconceptions that are hindering a student's progress, educators can provide targeted support and remediation, leading to more efficient and effective learning. While this study focuses primarily on cohort-level analysis, the foundational principles can be extended to individual student analysis in future work, potentially using LLMs to analyze responses to more qualitative, open-ended questions (as demonstrated in a limited way in this study's analysis of short-answer questions).

This study explores the potential of combining quantitative performance analysis with LLM-based characterization of misconceptions in online biomedical science courses. By employing classical test theory (CTT), along with validation via item response theory (IRT), to identify topics that are persistently challenging across cohorts, and then using LLMs to analyze student response patterns within these topics, we aim to provide a comprehensive understanding of not just *where* students struggle, but *why* (DeMars, 2010; Wilson, 2023). This mixed-methods approach offers insights that can inform both immediate teaching practices and long-term curriculum design.

Our methodology specifically focuses on quiz-level analysis rather than individual questions, examining student performance across 40-50 topic-focused quizzes per course. This approach allows us to identify topics that are both challenging and central to course objectives, creating a framework for prioritizing educational interventions. By analyzing first-attempt performance on multiple-choice questions (MCQs) across multiple cohorts, we identified topics that emerged as challenging across cohorts. This “consensus-based” approach revealed substantial overlap in difficult topics across cohorts, while allowing for potential variations between different course iterations.

This research addresses three primary questions:

- **Research question 1 (RQ1):** How can formative assessments be used to reliably identify specific topics where students struggle across different courses?

- **Research question 2 (RQ2):** To what extent can LLMs facilitate automated identification and characterization of misconceptions related to student difficulties in challenging topics?
- **Research question 3 (RQ3):** What are educators' perspectives on the potential impact of this data-driven approach to challenging topic identification for their teaching practice?

To answer these questions, we analyze data from a large cohort of medical students over three years, focusing on their performance in online courses covering fundamental biomedical sciences. Our methodology combined identification of challenging topics with characterization of misconceptions to provide a comprehensive analysis of topic difficulty. This analysis is supplemented by educator interviews to understand the practical implications of our findings.

The results of this study have implications for both course design and the broader application of artificial intelligence in educational settings. By providing a scalable, data-driven method for identifying and characterizing student misconceptions, this research offers educators valuable tools for enhancing teaching effectiveness and developing evidence-based interventions to address fundamental learning challenges. This work contributes to the growing body of research on the use of AI to personalize and improve learning, offering a practical approach for improving education in demanding fields like biomedical science.

# 2 Methods

## 2.1 Courses and students

We analyzed data from five online courses in the biomedical sciences, with 2,069 students from an international medical school in Mexico accounting for 3,802 course enrollments over the period studied (July 2019 to December 2023). Students were in Year 1 or Year 2 of their medical school curricula at the time of course enrollment.

This study includes 9 instances for 5 course types, each run over 10.5 weeks, with course content, including videos and assessment questions, released weekly, followed by a summative final exam. To reduce the impact of outliers and individual variability, course instances were grouped chronologically into cohorts, aiming for at least 200 enrollments per cohort per course type whenever possible. This threshold aligns with recommendations from psychometric research, where a minimum of 200 students per group helps stabilize item difficulty estimates in item response models (DeMars, 2010). Table 1 shows the breakdown of learners for each cohort over the study period.

**Table 1** Groupings (cohorts) with course dates and enrollment counts used in analysis for determining the hardest topics by cohort

| Course type (total enrollments) | Cohort 1 | Cohort 2 | Cohort 3 |
|---|---|---|---|
| Immunology (765) | July 2019, January 2020, July 2020<br>Total enrollments: 256 | January 2021, July 2021, January 2022<br>Total enrollments: 279 | July 2022, January 2023, July 2023<br>Total enrollments: 230 |
| Genetics (698) | August 2019, January 2020, July 2020<br>Total enrollments: 240 | January 2021, July 2021, January 2022<br>Total enrollments: 275 | July 2022, January 2023, July 2023<br>Total enrollments: 183 |
| Biochemistry (745) | July 2019, January 2020, July 2020<br>Total enrollments: 200 | January 2021, July 2021, January 2022<br>Total enrollments: 286 | July 2022, January 2023, July 2023<br>Total enrollments: 259 |
| Pharmacology (811) | October 2019, March 2020, October 2020<br>Total enrollments: 273 | April 2021, October 2021, April 2022<br>Total enrollments: 276 | October 2022, April 2023, October 2023<br>Total enrollments: 262 |
| Physiology (783) | September 2019, March 2020, October 2020<br>Total enrollments: 265 | April 2021, October 2021, April 2022<br>Total enrollments: 263 | October 2022, April 2023, October 2023<br>Total enrollments: 255 |

These specific online courses were selected because they provide a highly structured, data-rich environment with consistent assessment items across multiple cohorts, facilitating robust longitudinal analysis. At the collaborating medical school, these courses served as the students' primary initial exposure to the material, typically completed prior to the corresponding in-person coursework. While participation in these online modules was voluntary, the high enrollment numbers (Table 1) indicate they served as a key resource for student learning. The positioning as an initial, low-stakes learning opportunity before classroom instruction makes the data valuable for early identification of foundational misconceptions.

## 2.2 Videos and Assessment questions

All courses follow a similar format, with videos and a variety of visualizations interleaved with assessment questions, followed by a final exam. Assessments typically follow videos, allowing the learner to perform knowledge checks and incorporating spaced repetition and interleaving into the courses (Cepeda et al., 2006; Rohrer, 2012). All assessment questions in the lessons are considered formative assessment, as opposed to the final exam, which is considered summative assessment. Key features that make the assessment formative are not only the frequency and positioning of the assessment (interleaved with the content), but also the allowance of two attempts for each question and the provision of detailed explanations at the time each question is taken that address the correct and incorrect answers. This rapid feedback loop allows students to self-correct, identify knowledge gaps, and learn from their mistakes. Instructors are also provided with ongoing results and use those to design targeted interventions. For instance, instructors have used this information to address common mistakes in review sessions during a course. Finally, the assessment questions incorporated aspects of clinical and real-world relevance. Questions were designed to cover multiple levels of Bloom's taxonomy, with many requiring synthesis and integration, rather than factual recall.

Most of the formative assessment questions are in "key concepts" sections, where the core concepts are taught via concept videos (narrated, hand-drawn videos, each typically under 10 minutes in duration) with interleaved quizzes. Each quiz contains between 1 and 7 questions. The key concepts sections teach the main concepts of each topic, with the corresponding quizzes mapping directly to the main concepts from each course. This study therefore focused on the key concepts quizzes. The number of key concepts quizzes and quiz questions for each course type are shown in Table 2.

Most of the key concepts assessment questions are MCQs, including multi-select checkbox questions, with a smaller number of short answer questions, dropdown selection questions, and drag-and-drop questions (Table 3). Drag-and-drop questions were excluded from misconceptions analysis, given the difficulty of automatically characterizing these in a format that captured enough information about possible interactions for the LLM analysis.

**Table 2** The number of quizzes and number of formative assessment questions in the key concepts questions of each course

| Course type | Number of quizzes | Number of questions |
|---|---|---|
| Biochemistry | 39 | 108 |
| Genetics | 43 | 142 |
| Immunology | 42 | 139 |
| Pharmacology | 41 | 122 |
| Physiology | 39 | 85 |

**Table 3** Breakdown of course questions by type. Checkbox is a form of MCQ that allows for multi-select. Short answer questions allow freeform input.

| Question type | Percentage of questions |
|---|---|
| Multiple Choice | 69.1 |
| Checkbox | 15.9 |
| Short Answer | 8.3 |
| Dropdown | 6.2 |
| Drag and Drop | 0.5 |

Questions were designed based on best practices (for example, no "all of the above" type MCQs, plausible distractors, etc.) (Butler, 2018). All questions in the lessons allowed two attempts for mastery learning. Scores from the first attempts only were used in assessing the difficulty of each question; similarly, only the first attempt was used in determining the most common incorrect answer choices. First attempt results have been found to have the highest correlation with student outcomes in prior work, and we felt that these results would be most indicative of students' misconceptions(Parker et al., 2025)..

The number of key concepts questions attempted for each course is shown in Table 4.

**Table 4** Number of key concepts questions attempted per course

| Course type | Key concepts question attempts |
|---|---|
| Biochemistry | 62,292 |
| Genetics | 75,436 |
| Immunology | 90,090 |
| Pharmacology | 74,806 |
| Physiology | 48,949 |

## 2.3 Finding challenging topics

To identify the most challenging topics for students, we analyzed quiz performance data across three distinct cohorts (detailed in Table 1). The analysis proceeded as follows: First, for each course type, we calculated the average score of students' initial attempts on each key concepts quiz question. From these averages, the ten quizzes with the lowest average scores were selected for each cohort, representing the areas of greatest difficulty. Each selected quiz was then mapped to a corresponding topic heading (concept). This mapping was based on the title(s) of the lecture video(s) associated with each quiz. It's important to note that multiple quizzes could map to the same concept, particularly in cases where a concept was divided into multiple parts (e.g., "Concept X - Part 1" and "Concept X - Part 2"), each with its own associated quiz.

Following this initial identification, we performed a cross-cohort comparison to determine the most consistently challenging topics. We focused on the ten lowest-scoring topics for each cohort and applied two filtering criteria. First, a topic was considered consistently challenging only if it appeared among the ten most difficult topics in at least two of the three cohorts. Second, to ensure a reasonable sample size and mitigate the influence of potential outliers, we included only quizzes comprised of two or more questions. This excluded rare cases where a quiz consisted of only a single question. This filtering process yielded a final set of topics representing the areas of greatest and most consistent difficulty for students across the cohorts. The overall process is summarized in Fig. 2.

**Fig. 2** Flowchart of the process for identifying persistently challenging topics

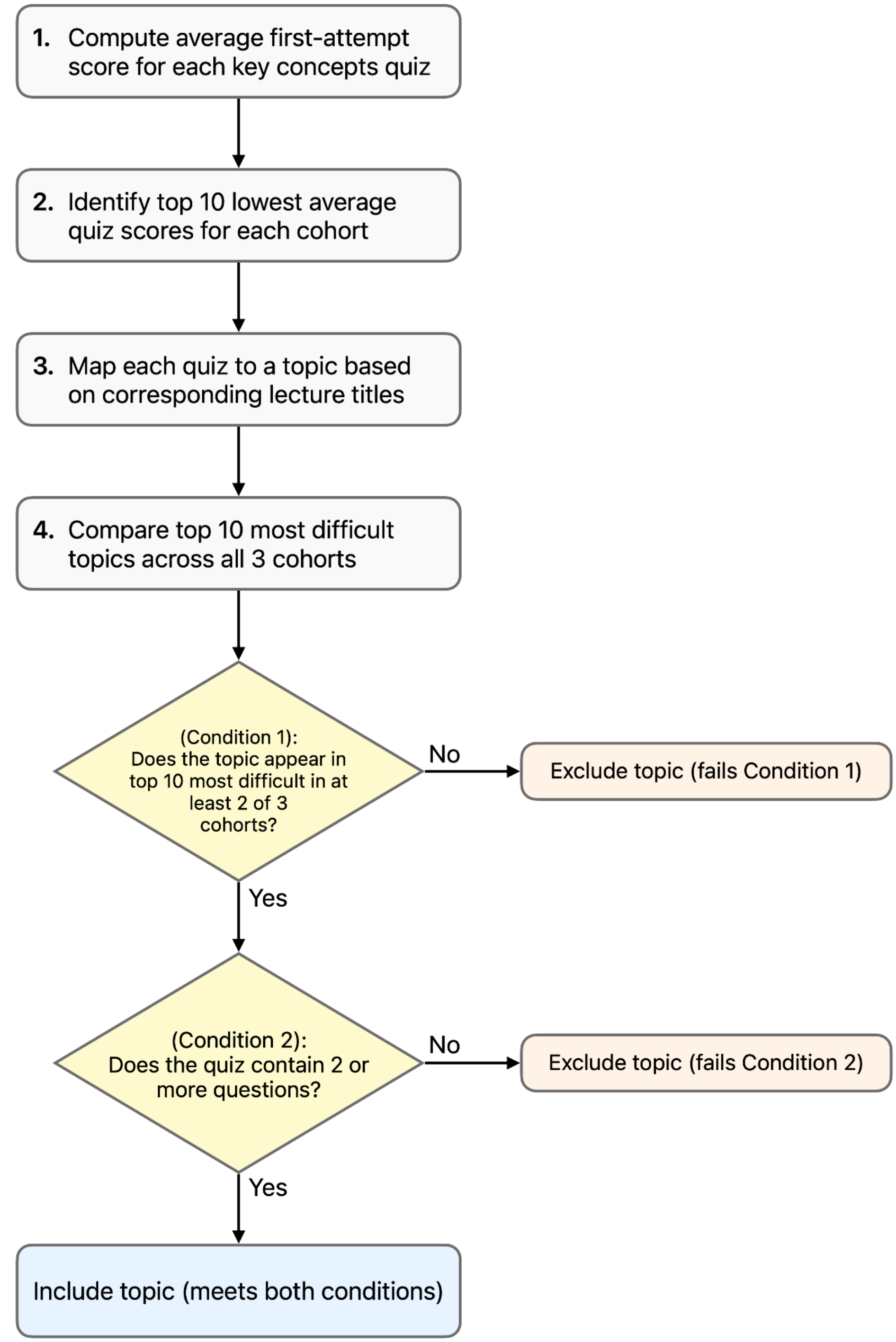

The ranking of the quizzes was also validated via IRT for one course type, Immunology, by using a 1-parameter logistic model (Rasch model). We used the py-irt package to find each question's difficulty based on the results from the 765 students taking the quiz questions across the three cohorts. For each quiz, the average item difficulty was then used to rank the quizzes and the Spearman correlation against the CTT ranking of those same quizzes was taken. The Spearman correlation was .99, $p < 10^{-5}$.

## 2.4 Creating Topic Synopses

The topic headings derived from the lecture titles were not descriptive enough (based on teacher feedback) to convey sufficient information to the teachers of the corresponding subjects in the medical school's curriculum. We therefore sought to create more complete topic descriptions using LLMs.

To do this, we prompted Gemini 2.0 Flash (at temperature 0 with all other model settings at default values) via the API with the topic synopsis prompt shown in the appendix, which combined the topic title and the video transcripts of the associated lecture videos (Fig. 3). Topic synopses were presented along with the challenging topics results in the interviews with professors.

**Fig. 3** Process for generating topic synopses. This flowchart shows the data sources (topic title, transcripts) and the LLM prompt structure used to generate a structured synopsis for course instructors. The full prompt is available in the appendix. We used Gemini 2.0 Flash for this step.

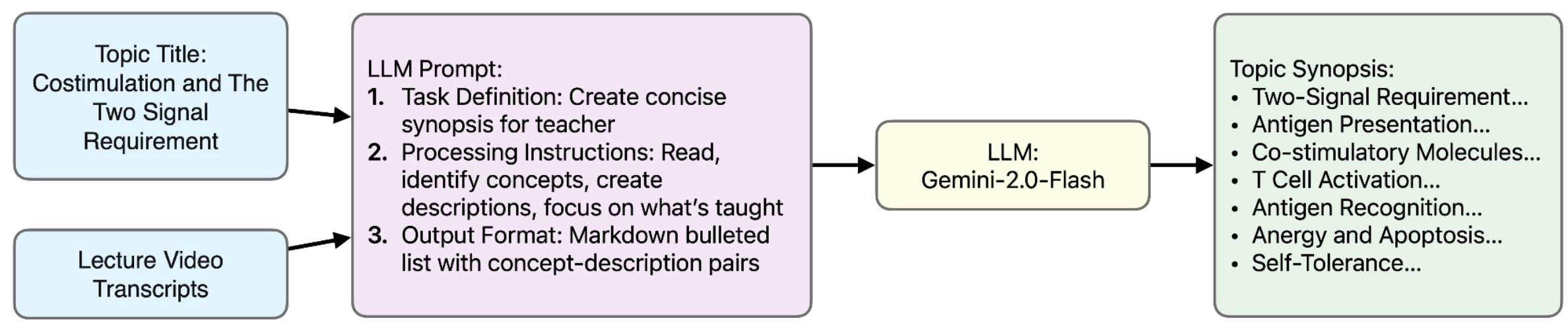

## 2.5 Interviews of Professors

To gain qualitative insights into the pedagogical context surrounding the challenging topics identified through quantitative analysis of formative assessment data, we conducted semi-structured interviews with faculty members at the collaborating Mexican medical school. These professors were selected based on their involvement in teaching the in-person courses corresponding to the online courses analyzed in this study. A total of ten professors were interviewed: two each for Immunology, Pharmacology, Physiology, Genetics, and Biochemistry.

The interviews were conducted in Spanish, the native language of the participants and interviewer, using a standardized, semi-structured interview protocol. Prior to the interview, participants were provided with a written description of the study's purpose and assured of the confidentiality of their responses. Verbal consent was obtained from all participants before commencing the recording. Transcripts of the interviews were translated into English using the Google Gemini-1.5-Pro AI model. To ensure translation accuracy, the English versions were cross-referenced and verified by the interviewer, a bilingual native Spanish speaker. This interviewer is a professor who was familiar with the curriculum and fluent in both medical terminology and academic discourse.

The interview protocol was divided into four main sections, the first three of which (described in the appendix) collected background information on the professor's background and teaching practice. The final section of the interview focused on the professors' perceptions of the challenging topics identified in the online courses. Interviewees were presented with a list of the top 10 most challenging topics (found via the methods described elsewhere in this paper) and a brief description of the content covered in those quizzes. Key questions in this section included:

- "What is your perspective on the potential of knowing these difficult topics that were found in the online courses?" (Exploring general utility)

- "Do you think this list of difficult topics can be used as an external tool to assess students' knowledge?" (Assessing validity as an external measure)
- "Do you think this list of difficult topics will help you to plan or change the way in which the topics are taught?"
- "How do you think the learning of these difficult topics can be improved?" (Exploring potential pedagogical changes)
- "Do you identify any kind of difficulty for this information to be taken precisely as an external evaluation?" (Identifying potential obstacles)

The semi-structured format allowed for follow-up questions and probes to clarify responses and explore emerging themes. Inductive thematic analysis was used, based on the qualitative data analysis framework from Terry et al. (2017). After reading the interview transcripts, the authors each coded the interviews to capture distinct concepts such as 'match with classroom observations,' 'abstract nature of concepts,' 'insufficient class time,' 'student lack of prior knowledge,' and 'utility for planning.' These codes were compared and grouped to identify the main themes reported in the results: Corroboration (validity of topics), Conceptual Complexity (intrinsic difficulty), and Pedagogical Context (structural factors). Passages corresponding to these themes were grouped together and are discussed in section 3.5.

## 2.6 Finding and Characterizing Misconceptions

To prepare the data for misconception identification by a large language model (LLM), we integrated several disparate data sources through a multi-stage process. This process involved data extraction, transformation, and the construction of the prompt for the LLM.

The quiz questions themselves were processed from their initial representation. XML representations of each question (the format that the questions were stored in on the learning platform, a version of Open edX) were parsed to extract a structured representation. This included the question stem, the available answer choices, the correct answer, the provided answer explanation (what is shown to the student after they submit their answer), and the question type (e.g., multiple choice, checkbox selection, dropdown, short answer, or drag and drop).

Student response data was derived from JSON log files. Because the primary database table only stored students' final attempts on each quiz question, it was necessary to analyze the log files to determine their initial attempts. From these initial attempts, we identified the most frequently selected incorrect answer choice for each question. This was achieved by calculating answer choice selection frequency for those first attempts.

Additionally, the content of the lecture videos associated with the quiz questions was incorporated. We employed the Whisper large-v2 model, running locally, to generate accurate transcriptions of the video lectures.

Finally, these three data sources – the structured question representation, the most common incorrect answer choice, and the relevant lecture transcript – were synthesized (Fig. 4) into a unified LLM prompt (see Fig. 8 in the Results section for a detailed visual example of this prompt structure and data synthesis). This prompt (exact format, with examples, shown in the appendix) was designed to provide the LLM with sufficient context to infer potential student misconceptions. Specifically, each prompt was processed using GPT-4 (the gpt-4o-2024-08-06 checkpoint, with a temperature of 0.3 and all other model settings at default values), which generated both a set of hypothesized misconceptions and the reasoning behind their identification. The structured output schema is also shown in the appendix.

**Fig. 4** Flowchart showing sources and output format, combined via the prompt (exact format, with examples, shown in the appendix), for the step of identifying possible student misconceptions. The gpt-4o model was used for this step.

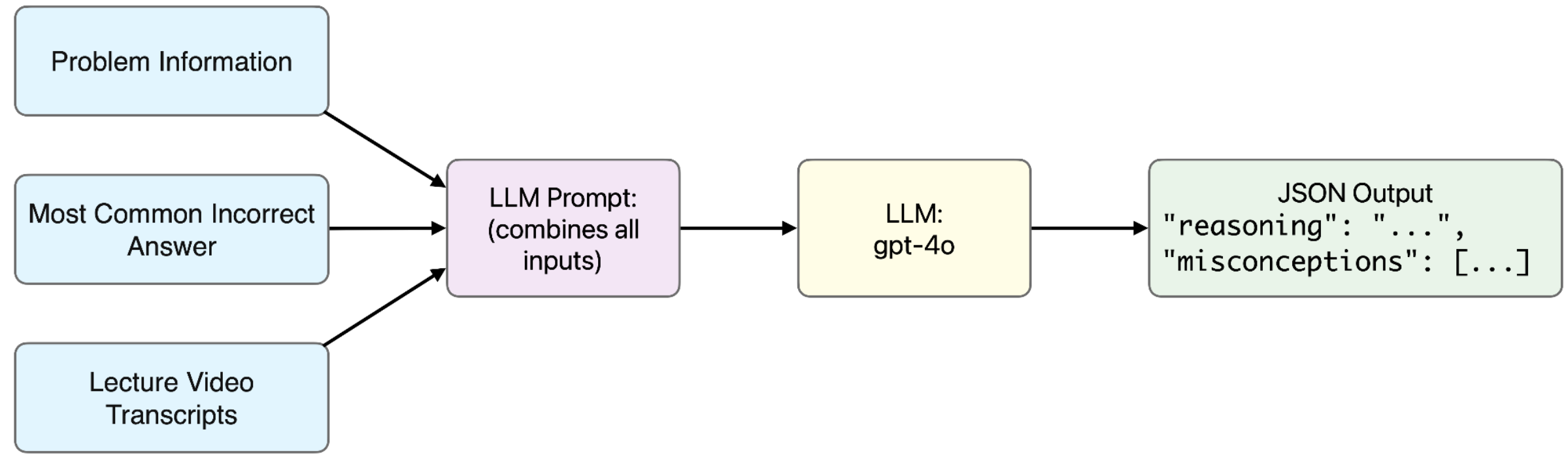

## 2.7 Evaluating Quality of LLM-Generated Misconceptions

To assess the validity and pedagogical utility of the LLM-generated misconceptions, we developed a structured evaluation rubric (see appendix). This rubric employed a 3-point scale (Excellent, Good, Poor) to rate each problem's misconception based on plausibility, immunological/physiological accuracy, relevance to the specific quiz question and incorrect answer choice, and clarity of expression. Raters could also optionally select a sub-code (A, R, C, or S) to pinpoint the primary area for improvement if a misconception was rated below "Excellent." This rubric was applied to the LLM output for two of the five courses: Immunology and Physiology. The Immunology misconceptions were evaluated by an immunology expert, and the Physiology misconceptions were evaluated by a physiology expert, both co-authors of this study. Each expert only evaluated misconceptions in their domain.

The expert evaluations focused on whether each proposed misconception represented a reasonable misunderstanding a student might hold, rather than attempting to identify the only possible misconception. This approach acknowledged the inherent complexity of student reasoning and prioritized the identification of plausible and actionable insights for instructors. The evaluation also served as a critical quality control step, ensuring that the LLM-generated content was not only logically consistent but also grounded in established scientific principles within the relevant domain. The evaluation provided data that could potentially be used to improve LLM performance via adjustments in prompts.

## 2.8 Data analysis and statistics

Data analysis was performed using a combination of Python libraries and API calls to large language models (LLMs). Statistical analysis and item response theory (IRT) modeling were conducted using py_irt (version 0.6.4), pandas (version 2.2.2) for data manipulation, and scipy (version 1.13.1) for statistics, including calculation of the Spearman correlation between the CTT and IRT results. To create transcriptions of lecture videos, we used the OpenAI Whisper large-v2 model. To identify challenging topics and misconceptions, we used two LLMs: OpenAI's gpt-4o-2024-08-06 and Google's gemini-flash-2.0, as mentioned above. Each model was used for a different part of the overall process: the gpt-4o model was used for identifying and characterizing the quiz question misconceptions (see section 2.6), and the gemini-2.0-flash model was used for creating topic synopses (see section 2.4). The gpt-4o model was chosen for the misconceptions analysis because a key differentiator of this model at the time the work was performed was its reliability in providing structured output. Access to these models was achieved through their respective APIs. Specific prompts used to query the LLMs, designed to elicit information about common student difficulties in biomedical science, are detailed in the appendix. Detailed methodology of processing LLM outputs can be found in other sections.

## 2.9 LLM cost and implementation considerations

The models were accessed using Python code written in Jupyter notebooks and/or Python scripts, requiring minimal infrastructure (e.g., an ordinary laptop). All LLM calls were performed via cloud-based endpoints (running on the model providers' infrastructure), with a cost for this project as shown below. The technical expertise required was primarily Python coding skills.

For reproducibility and scalability, calling the LLMs via API was important. Using these LLMs via their web-based chat interfaces, while possible, would be time-consuming and require significant manual effort for projects like this one where hundreds of LLM calls are needed. More importantly, the chat-based interfaces don't show certain parameters (for example, the model temperature), limiting reproducibility, and do not allow for reliable structured output.

The cost for LLM calls, shown in Tables 5 and 6, was minimal, with the misconceptions work requiring only approximately one dollar and the topic synopsis work requiring one cent (0.01 dollar) total, despite the inclusion of lengthy context from the video transcripts, problems, and explanations. (Tokens, the direct input for LLMs, are numerical representations of words or parts of words, with a token being approximately equal to three quarters of a word on average).

**Table 5** Finding and characterizing misconceptions using gpt-4o (input tokens: $2.50 per 1M tokens; output tokens: $10.00 per 1M tokens)

| Course Type | Input Tokens | Output Tokens | Total Tokens | Cost ($) |
|---|---|---|---|---|
| Biochemistry | 34,201 | 9,120 | 43,321 | 0.18 |
| Genetics | 54,326 | 12,160 | 66,486 | 0.26 |
| Immunology | 48,990 | 10,640 | 59,630 | 0.23 |
| Pharmacology | 45,640 | 13,680 | 59,320 | 0.25 |

| Physiology | 26,846 | 7,600 | 34,446 | 0.14 |
|---|---|---|---|---|
| Grand Total | 210,003 | 53,200 | 263,203 | 1.06 |

**Table 6** Creating topic synopses using gemini-2.0-flash (input tokens: $0.10 per 1M tokens; output tokens: $0.40 per 1M tokens)

| Course Type | Input Tokens | Output Tokens | Total Tokens | Cost ($) |
|---|---|---|---|---|
| Biochemistry | 10,940 | 1,606 | 12,546 | 0.0017 |
| Genetics | 14,783 | 1,992 | 16,775 | 0.0023 |
| Immunology | 14,391 | 1,704 | 16,095 | 0.0021 |
| Pharmacology | 13,523 | 2,491 | 16,014 | 0.0023 |
| Physiology | 10,003 | 1,565 | 11,568 | 0.0016 |
| Grand Total | 63,640 | 9,358 | 72,998 | 0.0101 |

# 3 Results

In this section, we organize the results in relation to the three research questions. The first research question pertains to whether formative assessments can reliably be used to identify specific topics where students struggle. The sections below on the *identification of challenging topics* and on *topic synopses via LLM*, in conjunction with the teacher interviews in the *teacher perspectives* section, answer this research question. The second research question explores whether LLMs can be used to automate finding misconceptions related to student difficulties on challenging topics. In the sections below on *finding and characterizing misconceptions via LLM* and *expert evaluation of LLM-generated misconceptions*, we demonstrate examples that show the potential for this use case and assess the quality of the models' output. The final research question asks about educators' perspectives on the potential impact of this approach to challenging topic identification to their teaching practice. The section below on *teacher perspectives* answers this research question.

Further description of the procedures used, along with the LLM models and parameters, can be found in section 2 (Methods).

We focus on the Immunology courses to show examples of results of the process, but the same process was performed for the other course types as well. Results for the other courses are shown in the appendix.

## 3.1 Identification of Challenging Topics

Following the procedure shown in Fig. 2, we analyzed quiz results across the 9 courses (3 cohorts) for each subject to derive the most challenging topics. Fig. 5 shows the specific process for Immunology, with more detailed results in subsequent figures/tables.

**Fig. 5** Steps for the challenging topic identification pipeline, with specific values shown for Immunology. The numbers were similar in scale for each course type but differed slightly. Each quiz corresponds to one topic.

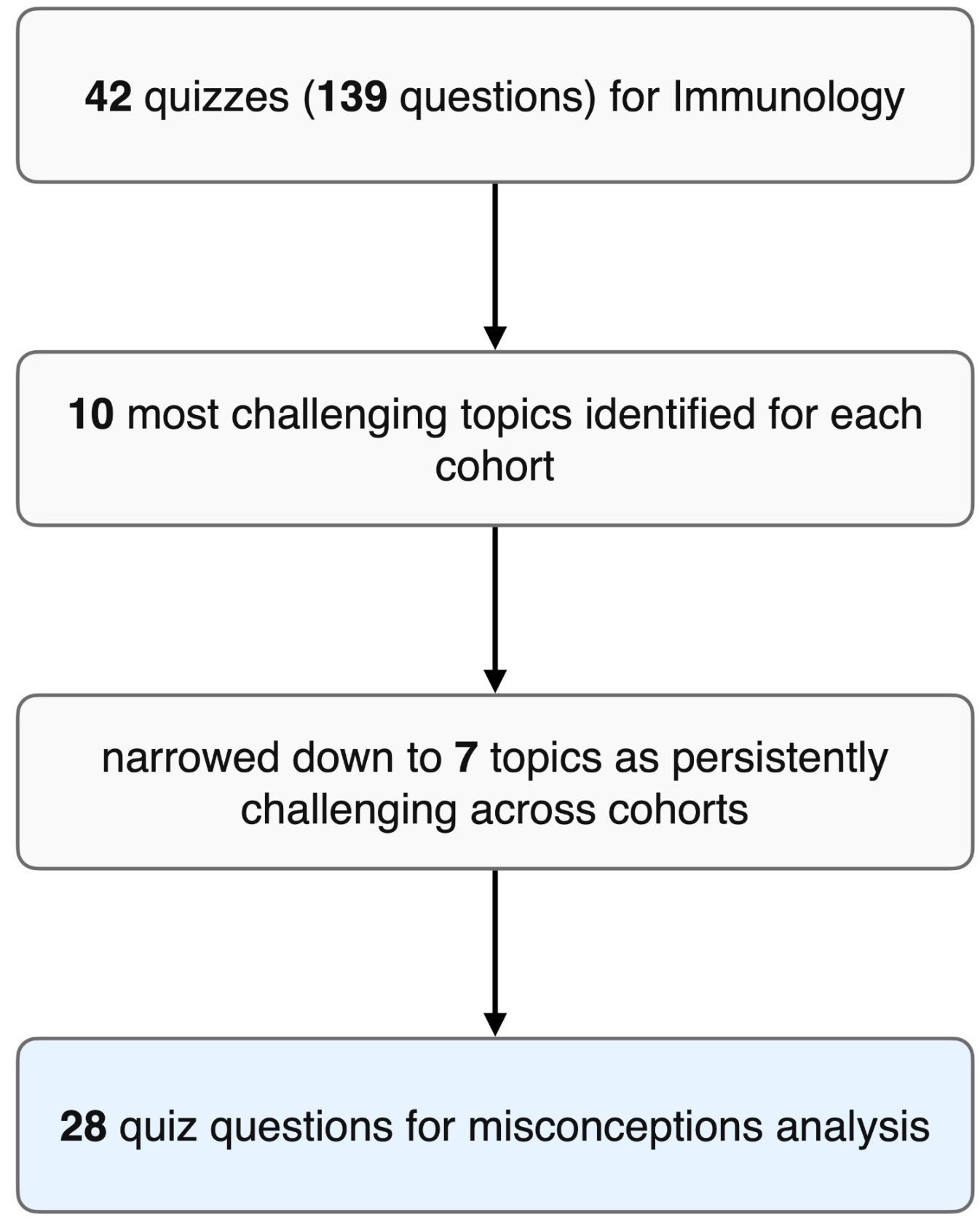

The difficult topics identified for each of the three Immunology cohorts, along with students' average quiz scores based on first attempts, are shown in Table 7. As noted in section 2.3, a quiz topic might show up more than once in this list if there was a part 1 and part 2 for that topic, each with a separate quiz (for example, see Novel Immune Therapies under Cohorts 1 and 2).

**Table 7** Topics identified as the top 10 most challenging for each of three immunology cohorts

| Cohort | Rank | Quiz topic | % Students correct on first attempt |
|---|---|---|---|
| Cohort 1 | 1 | Novel Immune Therapies | 50.9 |
| | 2 | The Germinal Center Reaction | 53.9 |
| | 3 | Costimulation and The Two Signal Requirement | 57.2 |
| | 4 | Novel Immune Therapies | 63.2 |
| | 5 | Immunological Barriers to Transplantation and Transfusion | 63.5 |
| | 6 | Response of Sentinel Cells | 63.6 |
| | 7 | How Do Antibodies Cause Disease? | 65.2 |
| | 8 | Immunodeficiencies | 67.4 |
| | 9 | Generation of Diversity | 67.8 |
| | 10 | Innate Recognition of Microbes | 68.0 |
| Cohort 2 | 1 | The Germinal Center Reaction | 47.0 |
| | 2 | Novel Immune Therapies | 48.8 |
| | 3 | Costimulation and The Two Signal Requirement | 57.3 |
| | 4 | Immunological Barriers to Transplantation and Transfusion | 58.5 |
| | 5 | Novel Immune Therapies | 58.8 |
| | 6 | Getting Around Barriers to Transplantation and Transfusion | 60.9 |
| | 7 | Response of Sentinel Cells | 62.6 |
| | 8 | Costimulation and The Two Signal Requirement | 63.0 |
| | 9 | Innate Recognition of Microbes | 63.0 |
| | 10 | Isotype Switching in the Germinal Center | 65.0 |
| Cohort 3 | 1 | Costimulation and The Two Signal Requirement | 60.6 |
| | 2 | The Germinal Center Reaction | 61.5 |
| | 3 | Immunological Barriers to Transplantation and Transfusion | 62.6 |
| | 4 | Novel Immune Therapies | 63.7 |
| | 5 | Innate Recognition of Microbes | 65.7 |
| | 6 | Response of Sentinel Cells | 65.8 |
| | 7 | T Cell-B Cell Collaboration | 68.7 |
| | 8 | Immunodeficiencies | 69.4 |
| | 9 | Costimulation and The Two Signal Requirement | 70.0 |
| | 10 | Introduction to Helper T Cells | 70.3 |

Fig. 6 shows this same information in a format that shows the support level (concurrence) across cohorts for each topic to be considered persistently challenging (see Fig. 2). Topics are included on the final list if they appear in the top 10 for two or more cohorts and their quizzes have 2 or more questions (no quizzes with only one question were present for this subject and were rare overall). The total number of quiz questions to be analyzed by an LLM for Immunology is shown in Table 8.

**Fig. 6** Each row denotes the top 10 challenging quizzes/topics for a given cohort. A '2' is shown in a box if there are two separate quizzes for that topic that are present in the top 10 for that cohort. The dashed box indicates the topics with appropriate support level to be considered persistently challenging.

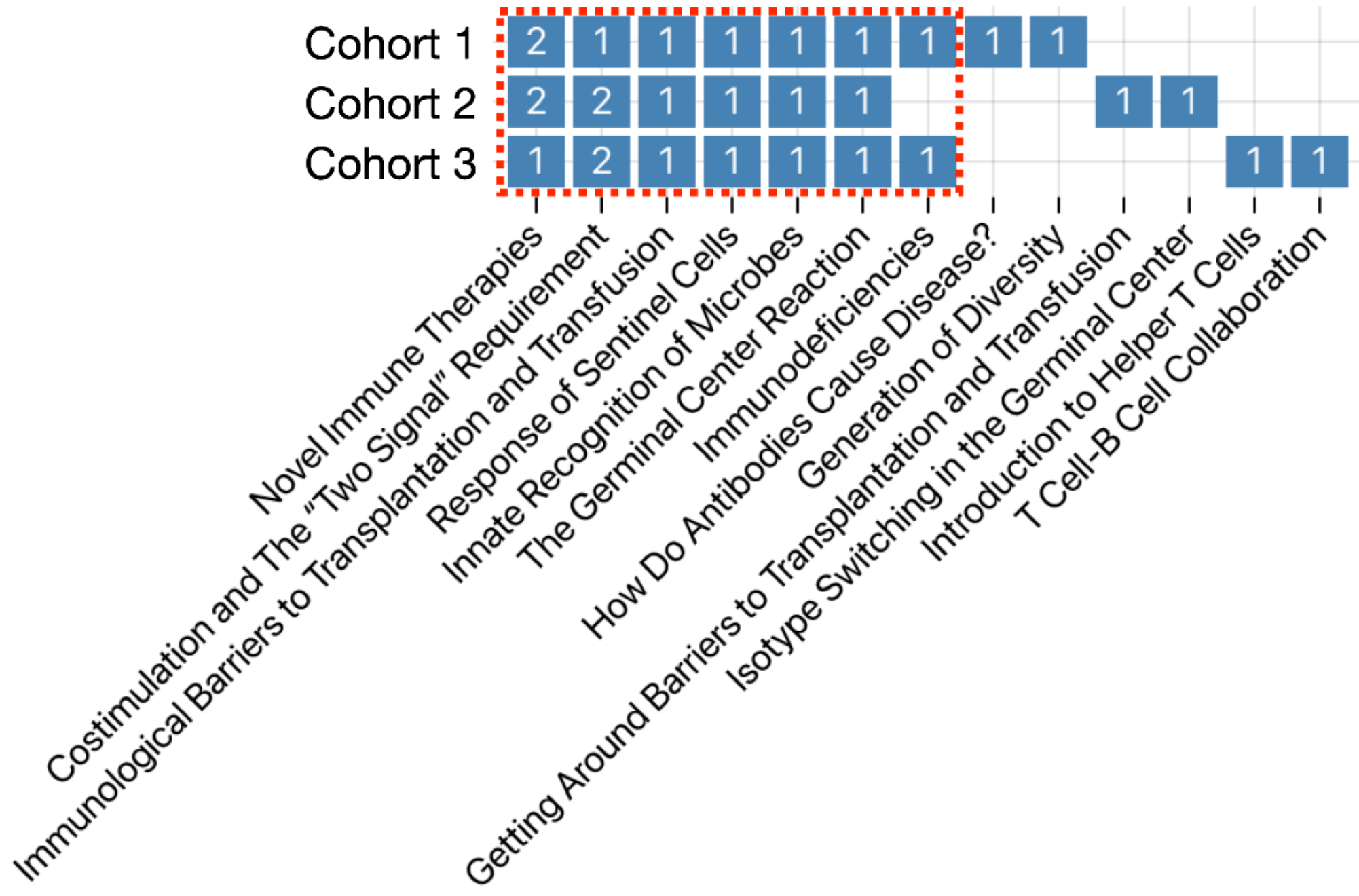

**Table 8** Final Immunology topics corresponding to the topics with sufficient support level in Fig. 6 and the number of questions for each topic quiz. There were 28 total quiz questions identified as persistently challenging for Immunology.

| Quiz topic | Cohort count | Number of quiz questions |
|---|---|---|
| Novel Immune Therapies | 3 | 5 |
| Costimulation and The “Two Signal” Requirement | 3 | 6 |
| Immunological Barriers to Transplantation and Transfusion | 3 | 4 |
| Response of Sentinel Cells | 3 | 2 |
| Innate Recognition of Microbes | 3 | 5 |
| The Germinal Center Reaction | 3 | 3 |
| Immunodeficiencies | 2 | 3 |

Across the five course types, the percent of identified topics that had appropriate support level ranged from 53% to 100% (full results shown in appendix Table 11). These results suggest that topics that are identified as highly challenging for one cohort are often, but not always, similarly challenging for other cohorts and that using support level criteria, as is done here, may form a useful prioritization step to separate cohort-specific challenges from persistent ones. Based on identification of these persistently challenging topics, we then used the quiz questions from those topics for further analysis to create topic synopses and to identify the specific misconceptions that students had on those topics’ quiz questions.

The challenging topics and topic synopses were then discussed with professors in interviews (described in the *teacher perspectives*) section in part to corroborate the findings of the analysis for research question 1.

## 3.2 Topic Synopses via LLM

The educators teaching the corresponding topics in the curriculum gave feedback that the quiz topics alone (presented with the identified misconceptions) were not descriptive enough to sufficiently understand what was included under those topics in the online courses. Although teachers had access to the online courses, rather than having the teachers review and characterize the lectures and other material themselves (a potentially time-consuming process), we sought to streamline the process for teachers by creating concise summaries for the coverage of each topic. The process is described in section 2.4 and involved providing an LLM with context that included the quiz topic and the transcripts from the associated lecture videos, along with specific instructions for output format and audience (see the appendix for the topic synopsis prompt). An example of the process applied to one of the Immunology quiz topics is shown in Fig. 7.

**Fig. 7** Example of the process for creating a topic synopsis for a topic in the Immunology course. The prompt is abbreviated for space reasons, with the complete version shown in an appendix. The LLM was called via the API.

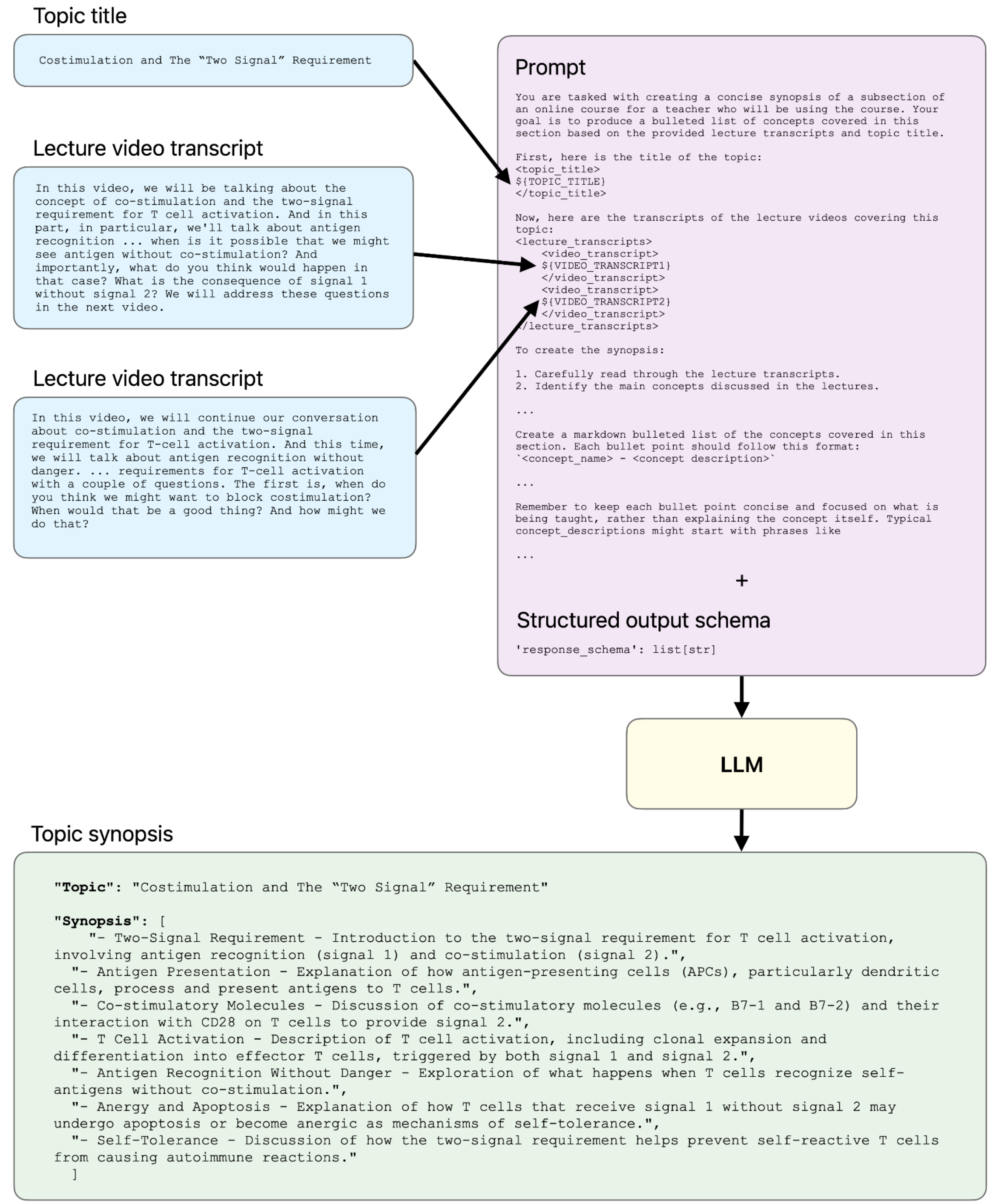

Given the audience of teachers, the model was instructed that it did not need to provide teaching explanations of the topics; rather, the goal was to concisely describe what was covered such that an immunology teacher would understand the specific aspects covered. These synopses were reported with the identified misconceptions described below.

## 3.3 Finding and Characterizing Misconceptions via LLM

In accordance with the procedure shown in Fig. 4, we used an LLM (GPT-4) to analyze each quiz question from the persistently challenging topic quizzes. As described in section 2.6, as data preparation, we 1) parsed quiz questions into an LLM-friendly format from the XML representations in which they were stored; 2) derived students' first attempts on the quiz questions from LMS log files; and 3) performed analysis to find the most common incorrect answer for each question. Along with the transcripts of the lecture videos, these formed the context given to the LLM via the prompt shown in the appendix.

An example for one Immunology problem is shown in Fig. 8.

**Fig. 8** The process by which context is combined and passed to the LLM for each problem, shown with one actual problem as an example. For space considerations, the lecture video transcripts and the system message are abbreviated. The complete system message is shown in the appendix. The LLM was called via the API.

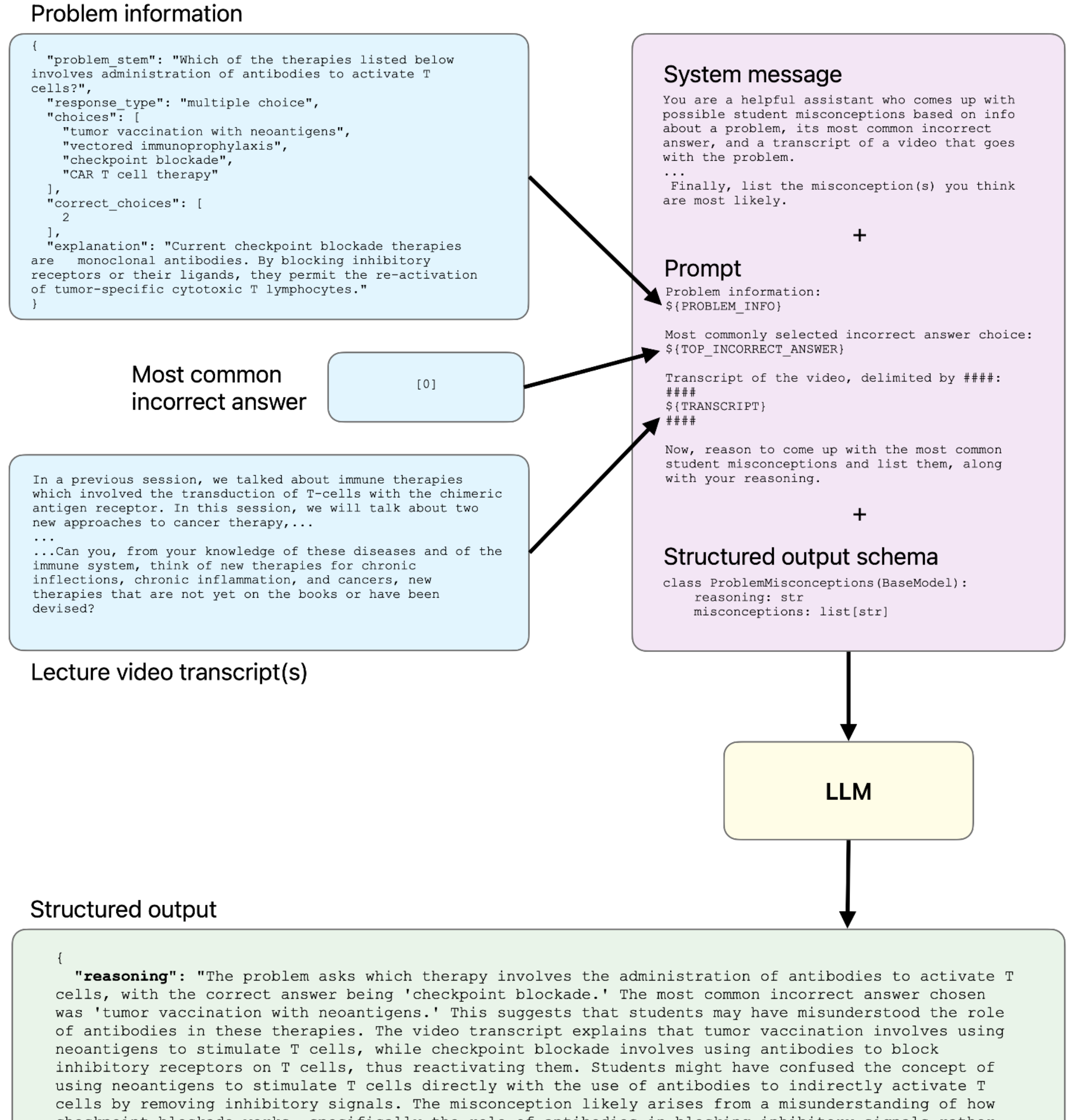
Problem information
{
"problem_stem": "Which of the therapies listed below involves administration of antibodies to activate T cells?",
"response_type": "multiple choice",
"choices": [
"tumor vaccination with neoantigens",
"vectored immunoprophylaxis",
"checkpoint blockade",
"CAR T cell therapy"
],
"correct_choices": [
2
],
"explanation": "Current checkpoint blockade therapies are monoclonal antibodies. By blocking inhibitory receptors or their ligands, they permit the re-activation of tumor-specific cytotoxic T lymphocytes."
}
Most common incorrect answer
[0]
In a previous session, we talked about immune therapies which involved the transduction of T-cells with the chimeric antigen receptor. In this session, we will talk about two new approaches to cancer therapy,...
...
...Can you, from your knowledge of these diseases and of the immune system, think of new therapies for chronic infections, chronic inflammation, and cancers, new therapies that are not yet on the books or have been devised?
Lecture video transcript(s)
System message
You are a helpful assistant who comes up with possible student misconceptions based on info about a problem, its most common incorrect answer, and a transcript of a video that goes with the problem.
...
Finally, list the misconception(s) you think are most likely.
+
Prompt
Problem information:
${PROBLEM_INFO}
Most commonly selected incorrect answer choice:
${TOP_INCORRECT_ANSWER}
Transcript of the video, delimited by ####:
####
${TRANSCRIPT}
####
Now, reason to come up with the most common student misconceptions and list them, along with your reasoning.
+
Structured output schema
class ProblemMisconceptions(BaseModel):
reasoning: str
misconceptions: list[str]
LLM
Structured output
{
"reasoning": "The problem asks which therapy involves the administration of antibodies to activate T cells, with the correct answer being 'checkpoint blockade.' The most common incorrect answer chosen was 'tumor vaccination with neoantigens.' This suggests that students may have misunderstood the role of antibodies in these therapies. The video transcript explains that tumor vaccination involves using neoantigens to stimulate T cells, while checkpoint blockade involves using antibodies to block inhibitory receptors on T cells, thus reactivating them. Students might have confused the concept of using neoantigens to stimulate T cells directly with the use of antibodies to indirectly activate T cells by removing inhibitory signals. The misconception likely arises from a misunderstanding of how checkpoint blockade works, specifically the role of antibodies in blocking inhibitory signals rather than directly stimulating T cells with antigens.",
"misconceptions": [
"Students may believe that tumor vaccination with neoantigens involves the administration of antibodies, confusing it with checkpoint blockade.",
"Students might not understand that checkpoint blockade specifically involves antibodies that block inhibitory receptors, leading to T cell activation.",
"There may be confusion between the direct stimulation of T cells by neoantigens and the indirect activation of T cells by removing inhibitory signals through antibodies."
]
}

For the example shown in Fig. 8, the reasoning and the identified misconceptions (with one example of the latter being, “There may be confusion between the direct stimulation of T cells by neoantigens and the indirect activation of T cells by removing inhibitory signals through antibodies.”) move beyond identifying surface-level recall issues. The reasoning (“*The video explains that checkpoint blockade uses antibodies to block inhibitory receptors on T cells, thus reactivating them. Students selecting 'tumor vaccination with neoantigens' might be confused about the role of antibodies in checkpoint blockade or might not fully understand the distinction between activating T cells through neoantigens and reactivating them by blocking inhibitory signals. They might also be conflating the concept of using antigens to stimulate an immune response with the use of antibodies to modulate that response, not recognizing that checkpoint blockade specifically involves antibodies to remove inhibition from T cells.*”) relates the potential misconception to what students heard in the instructional video, addressing the potential confusion between two different ways of activating T cells. This relationship between student answers and instructional material would not be apparent without providing the model with the content as context.

Another example, in this case from the Physiology course, involves a misconception identified for one of the acid-base problems. The misconception is “Buffers keep the concentration of the conjugate acid and conjugate base constant.” and the reasoning expands on this, including “*This misconception could stem from a misunderstanding of how buffers function chemically, possibly confusing the concept of equilibrium with the dynamic nature of buffering. The video transcript emphasizes that bicarbonate is not a buffer for respiratory acid, which might also confuse students about the role of bicarbonate in buffering systems, leading them to* *incorrectly apply the concept of constant concentrations to buffers in general.*" The ability of the model to examine the instructional content and hypothesize about why students are confused goes beyond simple identification of mistakes. Tying the misconceptions to the way the material is taught can also serve as useful feedback for instructors.

For each course type, we characterized misconceptions for all problems identified as part of persistently challenging topic quizzes. The number of problems processed for each course is shown in Table 9.

**Table 9** The number of problems identified from the challenging topics analysis for each subject. These problems were further analyzed to identify misconceptions.

| Course type | Number of problems on challenging topic quizzes |
|---|---|
| Pharmacology | 36 |
| Genetics | 32 |
| Immunology | 28 |
| Biochemistry | 24 |
| Physiology | 20 |

We then organized the problems and misconceptions the topic synopses into a comprehensive report for each subject that were then used as the basis for evaluation of the model output.

## 3.4 Expert Evaluation of LLM-Generated Misconceptions

To assess the pedagogical value and scientific accuracy of the LLM-generated misconceptions, expert faculty in immunology and physiology evaluated the LLM output using a structured rubric (detailed in the appendix). The rubric employed a three-point scale (Excellent, Good,

Poor) to rate the quality of each misconception based on its plausibility, accuracy, relevance to the question and incorrect answer, and clarity. If a misconception was rated below "Excellent," the evaluator optionally provided a code indicating the primary reason: Accuracy (A), Relevance (R), Clarity (C), or Specificity (S). This evaluation focused on identifying *plausible* misconceptions a student might hold, rather than exhaustively listing all possibilities.

Table 10 summarizes the expert evaluations for the immunology and physiology courses. These two courses were selected as representative of the broader set of five courses analyzed.

**Table 10** Expert Ratings of LLM-Generated Misconceptions

| Course | Problem type(s) | Total problems evaluated | Excellent (E) | Good (G) | Poor (P) | Breakdown of reasons for G/P ratings |
|---|---|---|---|---|---|---|
| Immunology | Mixed (MC, CB, SA) | 28 | 25 (89.3%) | 3 (10.7%) | 0 (0.0%) | G: 3S (Specificity); |
| Physiology | Mixed (MC, CB, SA, Dropdown) | 20 | 20 (100.0%) | 0 (0.0%) | 0 (0.0%) | |
| Combined | | 48 | 45 (93.8%) | 3 (6.2%) | 0 (0.0%) | |

*MC = Multiple Choice, CB = Checkbox, SA = Short Answer*

The vast majority of LLM-generated misconceptions were rated as "Excellent" (93.8% overall). This indicates that the LLMs were highly capable of generating plausible, accurate, relevant, and clearly stated misconceptions that would be useful for instructors to address.

- **Immunology:** Of the 28 immunology problems evaluated, 25 (89.3%) were rated "Excellent," 3 (10.7%) were rated "Good," and none were rated "Poor." The "Good" ratings were all attributed to a lack of specificity (S) to the incorrect answer choice. The added comments by the expert reviewer indicated specific places where student lack of reading instructions may be involved, whereas the model's identified misconceptions focused on concepts, as instructed in the prompt.

- **Physiology:** Of the 20 physiology problems evaluated, all 20 were rated "Excellent." The expert reviewer's comments identified instances where data presentation (missing bolded text in answer choices, a missing prior question, and obscured correct answer dropdown options) could be improved, and one instance of a subtle imprecision in the model's phrasing that did not affect the accuracy or quality of the identified misconceptions.

The high proportion of "Excellent" ratings, combined with the specific and feedback provided by the sub-codes for the "Good" ratings, suggests that LLMs provide a viable method for analyzing student misconceptions in undergraduate-level biomedical science courses. The detailed feedback from the rubric also offers valuable data for refining the prompting strategies used to elicit these misconceptions from the LLMs.

The expert evaluators did not observe any hallucinations, but it is important to note that that does not mean the model was perfect. For example, as noted above, one of the expert evaluators noted a small imprecision in the model's reasoning for one of the physiology questions, with no effect on the accuracy of the identified misconceptions. For that problem, the evaluator wrote "*All is correct on the misconceptions and accurate except for the last phrase of the reasoning ("leading them to think that the partial pressure of CO2 in the bottle returns to atmospheric levels") which is odd since the inside of the bottle was not at atmospheric levels to start. The rest is fine.*" Rigorous evaluation is still required to ensure the quality of output. The lack of hallucinations in this study suggests that providing relevant grounding/context is an effective strategy. We note this in the discussion as well.

## 3.5 Teacher Perspectives

To gain deeper insights into the pedagogical context surrounding the challenging topics identified through formative assessment data, we conducted semi-structured interviews with faculty members at the collaborating Mexican medical school. These interviews aimed to explore professors' perceptions of topic difficulty, potential underlying causes of student struggles, and the perceived utility of externally-derived data on challenging topics for informing their teaching practices. The thematic analysis of faculty interviews yielded three primary categories of findings: (1) Corroboration, where faculty confirmed the validity of the data-driven topics; (2) Conceptual Complexity, where faculty attributed student struggles to the inherent difficulty of the material; and (3) Pedagogical Context, where faculty identified structural challenges – such as time constraints and student preparedness – that contribute to learning difficulties. The results below are organized according to these themes.

A consistent theme across all interviews was the professors' general agreement with the identification of the challenging topics derived from the online course data. Professors from Biochemistry (Professor A), Genetics (Professors B and C), Immunology (Professors D and E), and Pharmacology (Professors F and G) largely confirmed that the topics identified as challenging in the online courses aligned with areas where they observed students struggling in their own classrooms. For instance, Professor A (Biochemistry) noted that glycolysis and beta-oxidation were frequent topics for which students requested additional tutoring, mirroring their low performance on related quizzes in the online courses. Similarly, Professor E (Immunology) concurred that innate immunity, adaptive immunity, and co-stimulation (topics flagged as challenging in the online course data) were consistently problematic for students, often requiring repeated explanations and alternative teaching approaches. Professor C (Genetics) stated that many students had difficulty in the molecular biology topics of transcription and translation, with which the online course data agreed. Professor F (Immunology) said "I agree with what I'm observing in my exam metrics that I'm doing, with what you are analyzing with exams, so I see a great advantage because one corroborates that they are complicated topics in immunology."

The interviews also provided valuable insights into the reasons behind student difficulties. Several professors highlighted the conceptual complexity inherent in many of the challenging topics. Professor B (Genetics) pointed out that topics like genomic imprinting and calculating risks in Mendelian or sex-linked inheritance often posed significant challenges due to their abstract nature and mathematical components. Professor D (Immunology) emphasized the need for students to understand the intricate interactions between receptors and cells in innate and adaptive immunity, a level of detail that often required repeated explanations and visual aids (e.g., drawing diagrams on the board). Professor A (Biochemistry) noted that some topics, such as $\Delta G$ (a change in energy that predicts how reactions proceed), were not primary teaching objectives, but essential context, leaving a gap. These comments support using student difficulty on formative assessment as an indicator of misconceptions, not just the difficulty of the assessment item.

Beyond conceptual complexity, several professors identified pedagogical factors that contributed to student difficulties. A recurring concern was the limited time available to cover complex topics adequately. Professor A (Biochemistry) noted that covering glycolysis in a single hour, for instance, made it challenging to delve into related pathways or provide sufficient context. Professor B (Genetics) concurred, stating that dedicating only one hour each to replication, transcription, and translation limited the depth of understanding. Professors also emphasized the transition from secondary education as a potential factor. Professor A pointed out that students entering medical school often lack well-developed study methodologies, making it harder to grasp complex topics in biochemistry. Professor C highlighted the mismatch between students' expectations of direct, basic questions and the applied, clinical-reasoning questions used in both the online courses and their own assessments.

These insights from faculty, particularly their agreement with the challenging topics identified through the formative assessment results, suggest that use of formative assessments is a viable way of automating identification of challenging topics.

# 4 Discussion

Addressing the critical need for scalable and insightful analysis of student learning difficulties, this study developed and validated a novel methodology integrating quantitative prioritization with LLM-based characterization of misconceptions. Traditional approaches often rely on educator intuition or methods requiring significant manual interpretation. This limits their applicability, especially in large cohorts (Gorgun & Botelho, 2023; Shi et al., 2021; Olde Bekkink et al., 2016). Our end-to-end approach overcomes these hurdles by first using quantitative performance data from formative assessments to identify persistently challenging topics and then employing course-context-enriched LLMs to generate plausible underlying misconceptions. This process emphasizes scalability, data-driven focus, and the production of contextually relevant misconceptions validated by experts.

Our **first research question** focused on the use of formative assessment results to find topics that are persistently challenging over time for students. Our findings strongly support the premise that formative assessments can reliably identify specific topics where students consistently struggle. Analyzing performance data from 3,802 student enrollments across multiple iterations of five distinct biomedical science courses provided a robust sample size. We found a high correlation (Spearman's $\rho = 0.99$, $p < 10^{-5}$ for Immunology) between difficulty rankings derived from classical test theory (CTT) and item response theory (IRT). This strong correlation validates our use of CTT-based difficulty metrics for prioritization.

Crucially, our focus on topic-level quizzes for the initial selection process, rather than individual questions, allowed us to identify broader conceptual difficulties, mitigating the noise from potential item-specific ambiguities or flaws. Furthermore, by requiring topics to be challenging across *multiple cohorts* (appearing in the top 10 most difficult for at least two out of three cohorts), we filtered out transient difficulties potentially caused by cohort-specific factors (e.g., the initial move to remote learning during the COVID-19 pandemic). The observation that, on average, approximately two-thirds (66%) of the most difficult topics were shared by at least two cohorts suggests that many challenges are persistent, likely stemming from the intrinsic complexity of the concepts or systematic gaps in the curriculum. This aligns with findings by Wiegand et al. (2021), who noted that identifying 'informatively hard' concepts is distinct from simple difficulty, though our approach automates this process using cross-cohort consistency. This cross-cohort consistency underscores the utility of formative assessment data for identifying high-priority areas for pedagogical intervention, enabling educators with limited time to focus their efforts where they are most needed.

Beyond identifying *where* students struggle, our **second research question** explored the feasibility of using LLMs to automate the understanding of *why*. The results demonstrate that current LLMs, when appropriately prompted and grounded, are highly capable in this complex task, even within a conceptually challenging area like undergraduate-level or early medical school biomedical science education. The minimal cost of using LLMs for identifying misconceptions (see section 2.9) and the ease-of-use via APIs make this a highly feasible process, even in resource-constrained settings. A key enabler was the provision of rich context to the LLM—for each question, including the quiz question itself, the correct answer, an expert explanation, common incorrect answer patterns, and relevant lecture transcripts. The ability of modern LLMs to process extensive context allows for a much more nuanced analysis, grounding the inference process in the specific learning materials students encountered and significantly mitigating the risk of LLM "hallucinations". Huang et al., (2025) list retrieval-augmented generation (providing additional retrieved knowledge, for example, relevant lecture transcripts, to the model along with the original query) as one of the strategies for hallucination mitigation. Prompt engineering techniques such as giving the model explicit instructions for cases where it has low confidence in its answer have also been used to mitigate hallucinations. Past research that provided only the quiz question as context focused on grade school or high school level topics that are likely to be more easily handled by the intrinsic knowledge of the LLMs (what was learned during the models' training). While Smart et al. (2024) demonstrated LLM utility for individual grade-school items, our results extend this validity to complex biomedical domains, suggesting that LLMs can function effectively even in specialized higher education contexts.

Techniques such as prompting the model to reason in detail before providing a final answer have been shown to increase output quality on complex tasks (Wei et al., 2022). In our case, we prompted for reasoning before the final list of misconceptions. Being able to show the reasoning to educators also gave greater insight as to why certain misconceptions were identified, potentially enhancing transparency and facilitating trust. Use of structured output provided a framework for consistency in output format over analysis of many quiz questions, including collecting the model's stated reasoning for identified misconceptions.

Human expert evaluations confirmed the high quality of the output, with 93.8% of identified misconceptions rated "Excellent" for plausibility, accuracy, relevance, and clarity. This strong performance suggests that highly capable LLMs (also called frontier models) possess sufficient subject matter expertise and reasoning capability, when combined with appropriate context, to analyze student errors effectively in complex domains. Generating concise topic synopses via LLM was a step that was added based on initial educator feedback; having this information also proved valuable, providing necessary context for educators reviewing the misconception analysis, highlighting that context is crucial for both the LLM and the human user.

Our **third research question** focused on educators' perspectives on the potential impact of this data-driven approach to challenging topic identification for their teaching practice. Interviews with faculty teaching corresponding in-person courses revealed strong alignment and perceived value. Professors overwhelmingly agreed that the topics identified as persistently challenging through the online course data mirrored difficulties they observed in their own classrooms. This "external validation" was highly valued by the professors, providing objective, data-driven evidence to corroborate experiential intuition and support potential curricular adjustments, such as incorporating more interactive elements or developing asynchronous resources.

Faculty expressed a clear desire for continued access to such data to dynamically tailor instruction and monitor the effectiveness of implemented changes. The faculty's desire for this data reinforces Ferguson et al.'s (2023) argument that learning analytics must be intelligible and actionable for instructors to effectively 'close the loop'. The synergy between the LLM-identified misconceptions and the qualitative insights provided by experienced faculty offers a powerful combination, augmenting teacher intuition with scalable, data-driven analysis to foster evidence-based pedagogical adjustments.

This work carries significant pedagogical implications:

- **Prioritized and Targeted Interventions:** By identifying persistently challenging topics and plausible underlying misconceptions, educators can prioritize and tailor interventions more effectively, focusing resources on areas with the greatest potential impact on student learning.
- **Feasibility and Scalability:** It presents a practical, evidence-based, and largely automated method for identifying and characterizing student misconceptions at scale—a task that would otherwise be prohibitively time-consuming for individual educators or curriculum teams.
- **Closing the Learning Analytics Loop:** This work facilitates the process by using data on student performance for analysis (identifying challenges and misconceptions), which informs targeted interventions (curricular changes, new resources, adjusted teaching strategies). Subsequent performance data can then be used to evaluate the effectiveness of these interventions, enabling a virtuous cycle of continuous improvement.
- **Foundation for Personalization:** While this study focused on cohort-level analysis, the principles are extensible to individual students. Analyzing individual response patterns across multiple questions, potentially combined with other data sources like observations from other types of learning activities, could inform personalized feedback, adaptive learning pathways, or individualized tutoring strategies.

As LLM capabilities advance, one can envision their use not only in identifying misconceptions but also in co-designing or even dynamically generating tailored learning activities or explanations to address specific student difficulties, subject to educator review and approval. For example, applying LLM analysis to a student’s answer to a multi-faceted, open-ended question could reveal nuanced misunderstandings. A “human-in-the-loop” paradigm would involve the LLM summarizing the student’s key misconceptions and providing recommendations for how a teacher might target those. In the future, a sufficiently capable LLM could use that information to design in real time an “N-of-1” intervention such as an exercise or interactive simulation specifically targeting that student’s misconceptions. A further example of a step toward personalization moves beyond individual quiz questions. An LLM could analyze the pattern of a student’s mistakes and misconceptions across the questions of an entire quiz or across multiple quizzes. This potentially unique pattern for each student could be analyzed relative to the pattern of the cohort, allowing each student to receive recommendations and feedback targeted at their unique set of misconceptions. “Reasoning” models (models trained with reinforcement learning and which use increased computing power to run queries) would likely be ideal both for the complex understanding of students’ misconceptions in advanced topics, as well as for developing interventions.

## 4.1 Ethical considerations and responsible AI use

Integrating AI into educational assessment, particularly in a high-stakes field like medical education, requires careful consideration of ethical implications such as transparency, bias, and responsible use. We designed our approach to address these concerns by using the LLM as a tool to support instructors, not to replace their judgment. To ensure transparency, we prompted the model to explain its reasoning, allowing educators to review its logic rather than trusting a 'black box.' This process builds trust and keeps the instructor in control. As mentioned elsewhere, grounding the LLM's analysis in specific course materials (lecture transcripts, quiz explanations from teachers, and student response patterns) helps mitigate the risk of biases stemming from the model's more general training data. Studies where detailed data about individual students are supplied as part of the information given to an LLM have a high potential for bias, depending on the scope of that data, and careful oversight is needed. The "human-in-the-loop" design, illustrated in Fig. 1, serves as an important safeguard. In our study, this involved faculty validating findings against their own expertise, a process demonstrated in our expert evaluations and teacher interviews. This ensures that the insights are used fairly and effectively to refine the curriculum, aligning the technology with a supportive and formative educational purpose. The path to further automation in the future requires trust developed through transparency and verification.

Although not specific to the use of AI, ethical use of learning data requires attention to appropriate data governance and proper consent for the use of student problem logs. For this study, IRB exemption for retrospective analysis of de-identified aggregate data was obtained (noted elsewhere in this paper). Furthermore, to strictly adhere to data governance regarding AI, we ensured that no individual student data or learning logs were transmitted to external LLM providers. The inputs provided to the model consisted solely of course content and aggregated statistics (the text of the single most common incorrect answer choice), ensuring student privacy was preserved.

Regarding linguistic bias in multilingual contexts, we acknowledge that translating non-English faculty interviews via AI can risk 'anglicizing' culturally specific pedagogical nuances or rhetorical styles. To mitigate this, we employed a human-in-the-loop verification process. As detailed in Section 2.5, AI-generated translations were cross-referenced and validated by the bilingual co-author who conducted the original interviews. This ensured that the analysis remained faithful to the participants' original intent and linguistic context, rather than relying solely on the model's training-data-dependent translation patterns.

# 5 Limitations

Several factors may influence the generalizability of these findings. First, our study focused on high-quality online biomedical science courses featuring numerous, well-structured, topic-based formative assessments developed with pedagogical expertise. The applicability of this method may be reduced in courses with significantly fewer assessment items, lower-quality questions, or assessments that do not accurately reflect the core learning objectives. Second, the study examined predominantly MCQs in quizzes that immediately followed instructional videos. MCQs constrain the space of possible responses and limit broader identification of misconceptions. An interesting avenue of investigation for future work involves studying the generalizability to other types of assessments such as open-ended responses or concept inventories and to delayed assessments (ones delivered weeks later, for example). Third, the research was conducted within the domain of biomedical sciences. While grounding with course-specific materials should enhance applicability across disciplines, generalizability to fields with less structured knowledge domains or different assessment paradigms (for example, assessment through open-ended discussions) requires further investigation.

We also acknowledge limitations specific to the use of LLMs, including hallucinations, dependence on training data, and contextual bias. Although strategies exist for mitigation of hallucinations (such as the grounding/RAG techniques discussed in section 4), hallucinations are not a solved problem, and educators should plan to perform verification checks. Furthermore, the lack of transparency in closed-source models regarding training data composition is a limitation; it remains unclear if the training data includes specific niche domains in sufficient depth. Educators should consult model benchmarks to understand performance strengths in their specific fields.

Regarding contextual bias, while our zero-shot approach relied on grounding the model with lecture transcripts to reduce hallucinations, the prompt structure itself introduces potential biases. By explicitly providing the most common incorrect answer in the prompt along with the associated lecture transcripts, we introduce an anchoring bias. The model may be primed to rationalize that specific error using the lecture material. While this approach benefits the analysis by grounding the model in the same context available to students, it may also lead to overlooking other valid student misconceptions, including those possessed by a minority of students. Future work could mitigate this bias using in-context learning. By including balanced few-shot examples demonstrating both correct reasoning and various types of incorrect reasoning, future implementations could better calibrate the model to distinguish between genuine misconceptions and the provided anchors.

Finally, the LLM was specifically prompted to identify *misconceptions*. This focus might sometimes overlook instances where student errors stem primarily from unclear question wording or formatting issues, as noted by one expert evaluator. Since these types of errors were not the focus of this study, the evaluation rubric could be refined to better distinguish them from conceptual misconceptions. This clarification of the rubric would be particularly helpful for evaluators for whom English is not their native language.

# 6 Concluding Remarks

This study demonstrates the power of systematically combining quantitative analysis of formative assessment data with the advanced capabilities of large language models to gain deeper insights into student learning challenges. We presented and validated a scalable methodology for identifying persistently difficult topics in online courses and characterizing the likely underlying student misconceptions associated with them. This approach moves beyond anecdotal evidence or manual qualitative analysis, offering educators objective, actionable insights grounded in student performance data and course context. The increasing sophistication of LLMs, including emerging multimodal capabilities and enhanced reasoning abilities, opens exciting avenues for extending this work to incorporate visual data (for example, quiz questions with images or interactivity), analyze lecture content directly (for example, directly giving video or audio to the model), and potentially improve accuracy across diverse and complex domains. Ultimately, this methodology provides a robust framework for leveraging AI to enhance teaching effectiveness and foster more targeted, evidence-based, and potentially personalized learning experiences.

# Appendix A

## Rubric for evaluation of misconception output

The rubric shown below is for Immunology. This was slightly adapted (changing the references to immunology only) for the other subjects.

**Instructions for evaluating misconceptions**

You are evaluating the quality of AI-generated student misconceptions in immunology. For each question, you will be given the question, answer choices, a topic summary, and a potential misconception a student might have based on choosing a specific incorrect answer.

Please rate each misconception using the following 3-point scale:

**3 - Excellent**: The misconception is highly plausible, clearly stated, immunologically accurate, and directly addresses the specific error highlighted by the incorrect answer choice. It would be very useful for an instructor to address.

**2 - Good**: The misconception is plausible and generally accurate but may be slightly vague, not entirely specific to the question, or have a minor immunological imprecision that doesn't fundamentally mislead. It's still useful but might need some refinement.

**1 - Poor**: The misconception is implausible, inaccurate, irrelevant to the question, or fundamentally flawed from an immunological perspective. It would not be useful or might even be misleading.

Focus on whether the misconception is plausible (could a student reasonably hold this belief?), immunologically accurate, relevant to the question and incorrect answer, and clearly stated. A 'good' misconception doesn't have to be the only possible misconception, just a reasonable one.

If you rate a misconception as anything other than 'Excellent (3)', please optionally choose ONE of the following codes to indicate the primary reason:

A - Accuracy Issue: The core immunological concept is incorrect or misrepresented.

R - Relevance Issue: The misconception doesn't seem connected to the specific question or the chosen incorrect answer.

C - Clarity Issue: The misconception is poorly worded, vague, or difficult to understand.

S - Specificity Issue: The misconception is too general and doesn't explain why a student might choose that particular wrong answer.

*Example*: A rating of "2-S" would mean "Good, but lacks specificity to the incorrect answer." A rating of "3" would mean no sub-category is needed.

Your expertise is invaluable in determining if these misconceptions are accurate and helpful for improving student understanding.

## Prompts

### Topic Synopsis prompt

```
You are tasked with creating a concise synopsis of a subsection of an online
course for a teacher who will be using the course. Your goal is to produce a
bulleted list of concepts covered in this section based on the provided
lecture transcripts and topic title.

First, here is the title of the topic:
<topic_title>
${TOPIC_TITLE}
</topic_title>

Now, here are the transcripts of the lecture videos covering this topic:
<lecture_transcripts>
${VIDEO_TRANSCRIPTS}
</lecture_transcripts>

To create the synopsis:

1. Carefully read through the lecture transcripts.
2. Identify the main concepts discussed in the lectures.
3. For each concept, create a brief, one-sentence description that captures
its essence.
4. Remember that the teacher already knows the subject material, so focus on
describing what is being taught rather than explaining the concept in detail.

Create a markdown bulleted list of the concepts covered in this section. Each
bullet point should follow this format:
`<concept_name> - <concept description>`

Where:
- <concept_name> is a short, clear name for the concept
- <concept description> is a single sentence describing what is being taught
about this concept
```

Your final output should be a concise list of 5-10 bullet points, depending on the complexity and breadth of the topic. Present your list inside <synopsis> tags.

```
Example format:
<synopsis>
- Concept A - Description of what is taught about Concept A.
- Concept B - Description of what is taught about Concept B.
- Concept C - Description of what is taught about Concept C.
</synopsis>
```

Remember to keep each bullet point concise and focused on what is being taught, rather than explaining the concept itself.

**Misconceptions Prompt**

System message:

```
You are a helpful assistant who comes up with possible student misconceptions
based on info about a problem, its most common incorrect answer, and a
transcript of a video that goes with the problem.

You will be given the problem information, the most common incorrect answer,
and the transcript of the video. First, carefully read all of the information
given to you. Then, reason through what you think the most common student
misconception(s) might be. Record your reasoning. Finally, list the
misconception(s) you think are most likely.
```

User message (the placeholders represent information that is filled in for each problem for the API call):

```
Problem information:
${PROBLEM_INFO}

Most commonly selected incorrect answer choice (0-indexed on the choices if
this is a list of integers):
${TOP_INCORRECT_ANSWER}

Transcript of the video, delimited by ####:
####
${TRANSCRIPT}
####

Now, reason to come up with the most common student misconceptions and list
them, along with your reasoning.
```

An example of the formatting of the PROBLEM_INFO variable:

```
{
```

```
  "problem_stem": "Which of the therapies listed below involves
administration of antibodies to activate T cells?",
  "response_type": "multiple choice",
  "choices": [
    "tumor vaccination with neoantigens",
    "vectored immunoprophylaxis",
    "checkpoint blockade",
    "CAR T cell therapy"
  ],
  "correct_choices": [
    2
  ]
}
```

An example of the formatting of the TOP_INCORRECT_ANSWER variable:

```
[0]
```

The transcripts of videos (example not included here for brevity) were the speaker text of the transcripts without paragraph breaks, time codes, or other formatting.

Misconceptions prompt structured output Pydantic schema:

```
class ProblemMisconceptions(BaseModel):
    reasoning: str
    misconceptions: list[str]
```

## Professor interviews process

The interview protocol was divided into four main sections. First, general information about the professor's teaching background was collected, including the subjects and programs taught, years of teaching experience, highest academic degree, and area of expertise. Second, the interview explored the professors' current practices for identifying challenging topics within their courses, including methods used, alignment with the online course-identified topics, time allocation to challenging topics, and typical teaching activities. Third, questions addressed the professors' use of electronic tools and their involvement in online course design and management (e.g., using the Canvas learning management system). The final section related to key questions surrounding the identified challenging topics and is described in the article.

## Challenging topic results from courses

**Fig. 9** Persistently challenging topics for Biochemistry. Each row denotes the top 10 challenging quizzes/topics for a given cohort. A ‘2’ is shown in a box if there are two separate quizzes for that topic that are present in the top 10 for that cohort. The dashed box indicates the topics with appropriate support level to be considered persistently challenging.

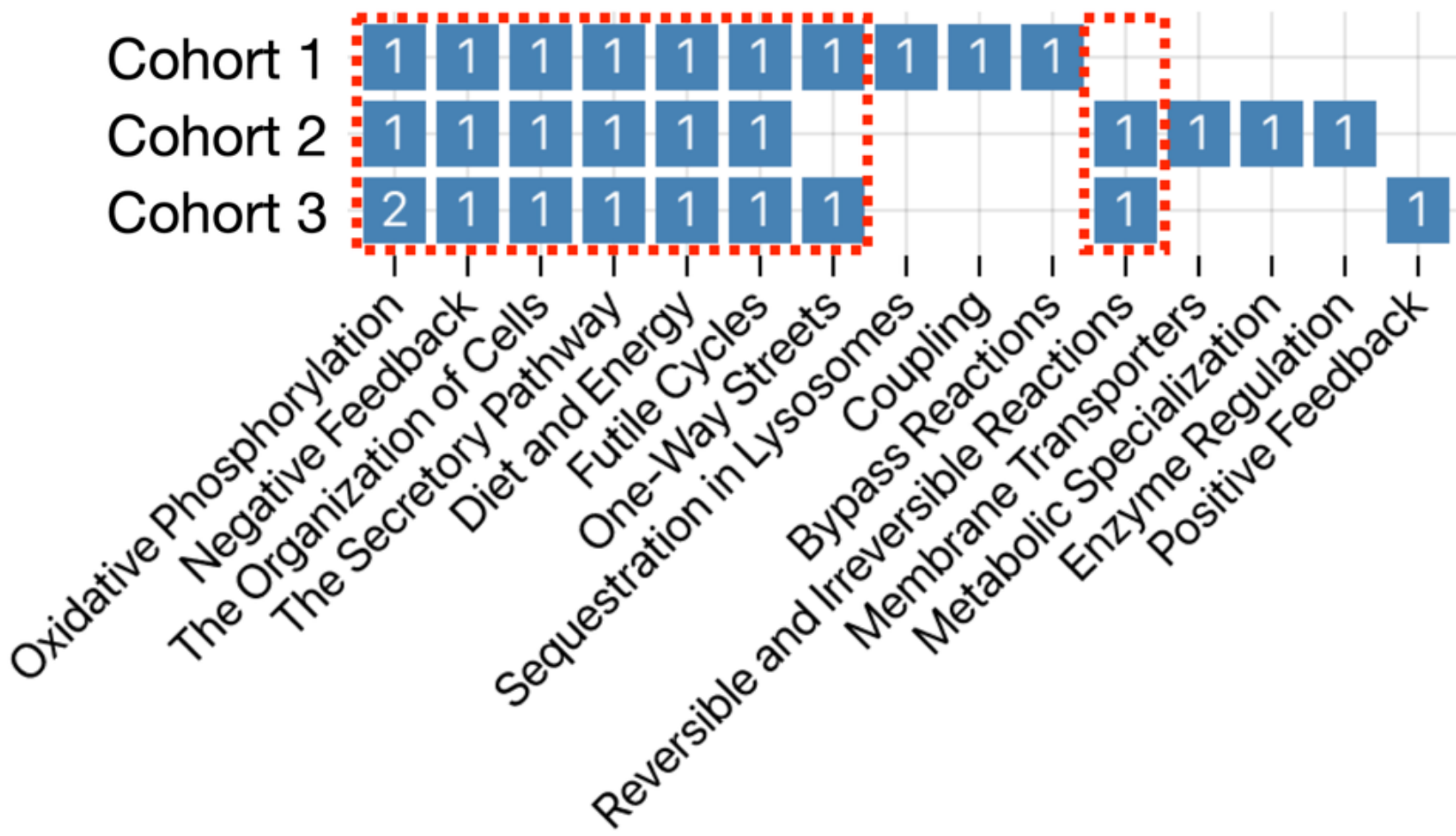

**Fig. 10** Persistently challenging topics for Genetics. Each row denotes the top 10 challenging quizzes/topics for a given cohort. A '2' is shown in a box if there are two separate quizzes for that topic that are present in the top 10 for that cohort. The dashed box indicates the topics with appropriate support level to be considered persistently challenging.

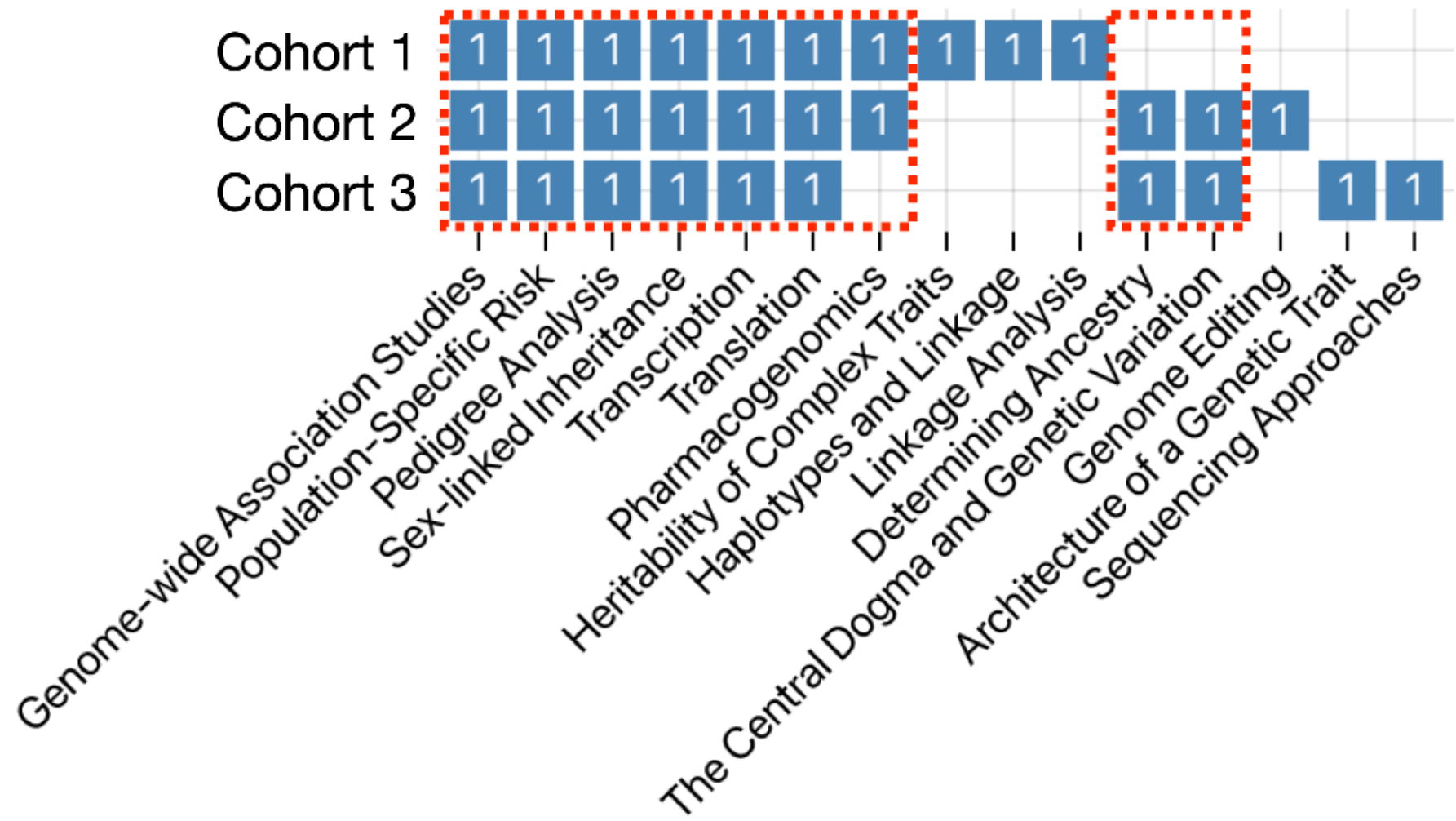

**Fig. 11** Persistently challenging topics for Pharmacology. Each row denotes the top 10 challenging quizzes/topics for a given cohort. A '2' is shown in a box if there are two separate quizzes for that topic that are present in the top 10 for that cohort. The dashed box indicates the topics with appropriate support level to be considered persistently challenging.

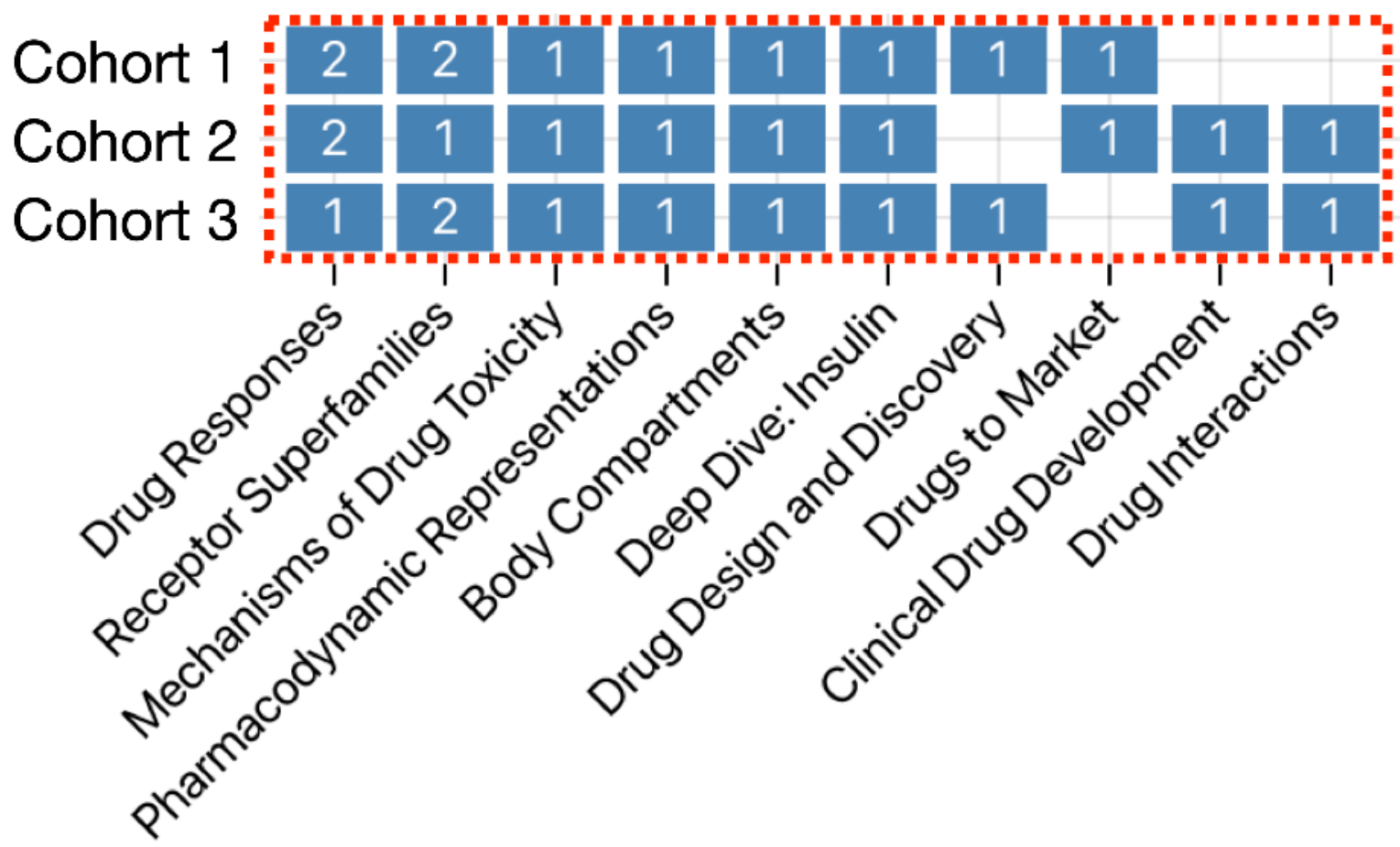

**Fig. 12** Persistently challenging topics for Physiology. Each row denotes the top 10 challenging quizzes/topics for a given cohort. A '2' is shown in a box if there are two separate quizzes for that topic that are present in the top 10 for that cohort. The dashed box indicates the topics with appropriate support level to be considered persistently challenging.

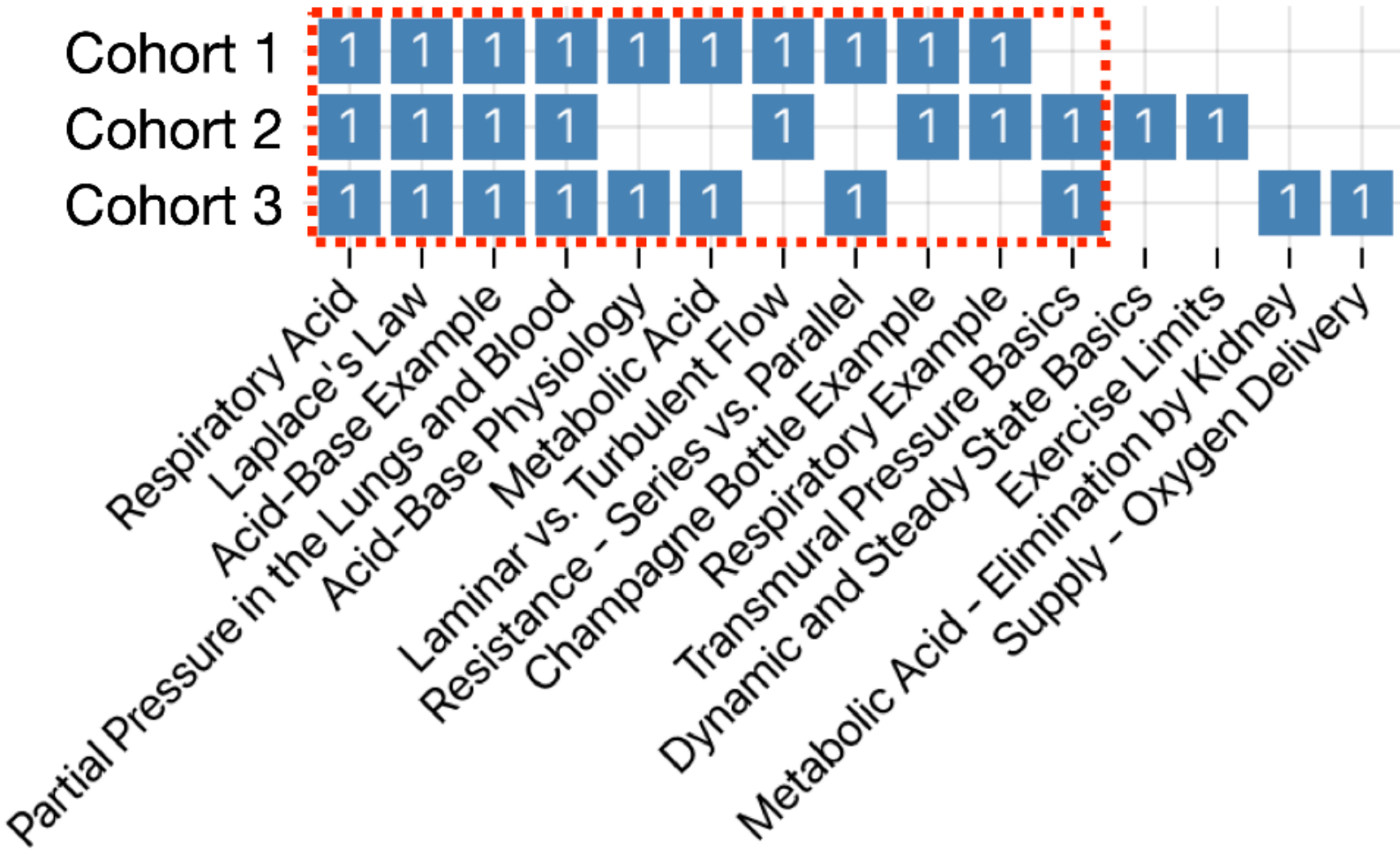

**Table 11** Number of quizzes identified as challenging and the number (percent) identified as persistently challenging across cohorts based on the support level criteria

| Course type | Number of quizzes identified as challenging (3 cohorts) | Number of quizzes identified as persistently challenging by support level criteria (%) |
|---|---|---|
| Immunology | 13 | 7 (54) |
| Genetics | 15 | 9 (60) |
| Biochemistry | 15 | 8 (53) |
| Pharmacology | 10 | 10 (100) |
| Physiology | 15 | 11 (73) |